\RequirePackage{etoolbox}
\newtoggle{arxiv}
\toggletrue{arxiv}

\newtoggle{final}
\iftoggle{arxiv}{%
    \documentclass[a4paper,11pt]{article}
    \toggletrue{final}
    \usepackage{arxiv}
    \usepackage[numbers,sort]{natbib}
    \newcommand{\cvspace}[1]{}
}{}
\usepackage{url}
\usepackage{booktabs}
\usepackage{amsfonts}
\usepackage{makecell}
\usepackage{rotating}
\usepackage{nicefrac}
\usepackage{microtype}
\usepackage{mathtools,bm,amssymb}
\usepackage{subcaption,placeins}
\usepackage{adjustbox,multirow,dcolumn}
\usepackage{algorithm,algpseudocode,setspace}
\usepackage{color}
\usepackage{graphicx}
\usepackage[most]{tcolorbox}
\usepackage[nomessages]{fp}
\tcbuselibrary{skins,breakable}
\usetikzlibrary{shadings,shadows}

\graphicspath{{figures/}}

\makeatletter

\newcolumntype{d}{D{.}{.}{3.2}}
\makeatletter
\newcolumntype{B}{>{\boldmath\DC@{.}{.}{3.2}}c<{\DC@end}}
\makeatother

\newcommand{\theader}[1]{\multicolumn{1}{c}{\textbf{#1}}}

\newcommand{\tpm}[2]{\( {\textrm{#1}}{\scriptstyle \pm\textrm{#2}} \)}
\newcommand{\tbpm}[2]{\tpm{\textbf{#1}}{#2}}
\newcommand{\tupm}[2]{\tpm{\underline{#1}}{#2}}
\newcommand{\tgpm}[2]{\textcolor{darkgray}{\tpm{#1}{#2}}}

\newcommand*\samethanks[1][\value{footnote}]{%
    \footnotemark[#1]\hspace{0.4em}}

\makeatletter

\DeclareRobustCommand\onedot{\futurelet\@let@token\@onedot}
\def\@onedot{\ifx\@let@token.\else.\null\fi}
\iftoggle{arxiv}{%
    \newcommand{\eg}{\emph{e.g\@\onedot}}
    
    \newcommand{\etal}{\emph{et~al\@\onedot}}
    \newcommand{\ie}{\emph{i.e\@\onedot}}

}{}

\iftoggle{arxiv}{%
    \newcommand{\iid}{\emph{i.i.d.}}
}{}
\newcommand{\noniid}{non-\iid{}}
\newcommand{\name}{\textsc{Flip}}
\iftoggle{final}{
    \newcommand{\sourcelink}{\url{https://github.com/0-ml/flip}}
}{
    \newcommand{\sourcelink}{[link anonymized for review]}
}
\newcommand{\fed}[1]{\textit{f}-{#1}}
\newcommand{\eval}[1]{\FPeval{\result}{clip({#1})}\result}
\newcommand{\numtrials}{3}
\newcommand{\numalgorithms}{8}
\newcommand{\numprotocols}{4}
\newcommand{\numscenarios}{6}
\newcommand{\numdatasets}{12}
\newcommand{\numglobaldatasets}{8}

\newcommand{\commcost}{\kappa}
\newcommand{\accglobal}{\alpha_\mathrm{g}}
\newcommand{\accpersonal}{\alpha_\mathrm{p}}
\newcommand{\accbase}{\alpha_\mathrm{b}}
\newcommand{\accnovel}{\alpha_\mathrm{n}}
\newcommand{\accharmonic}{\alpha_\mathrm{h}}
\newcommand{\accfewshot}[1]{\alpha_\mathrm{s=#1}}
\newcommand{\accxd}[2]{\alpha_\mathrm{\textrm{#1}\rightarrow\textrm{#2}}}
\newcommand{\textcommcost}{\( \commcost \)}
\newcommand{\textaccglobal}{\( \accglobal \)}
\newcommand{\textaccpersonal}{\( \accpersonal \)}
\newcommand{\textaccbase}{\( \accbase \)}
\newcommand{\textaccnovel}{\( \accnovel \)}
\newcommand{\textaccharmonic}{\( \accharmonic \)}
\newcommand{\textaccfewshot}[1]{\( \accfewshot{#1} \)}
\newcommand{\textaccxd}[2]{\( \accxd{#1}{#2} \)}
\newcommand{\txtg}[1]{\textcolor{gray}{#1}}

\DeclarePairedDelimiter{\parens}{\lparen}{\rparen}

\DeclarePairedDelimiter{\braces}{\{}{\}}

\DeclareMathOperator{\similarity}{{sim}}
\newcommand{\subalign}[1]{%
    \vcenter{%
        \Let@ \restore@math@cr \default@tag
        \baselineskip\fontdimen10 \scriptfont\tw@
        \advance\baselineskip\fontdimen12 \scriptfont\tw@
        \lineskip\thr@@\fontdimen8 \scriptfont\thr@@
        \lineskiplimit\lineskip
        \ialign{\hfil$\m@th\scriptstyle##$&$\m@th\scriptstyle{}##$\hfil\crcr
            #1\crcr
        }%
    }%
}

\makeatother

\iftoggle{arxiv}{\usepackage{cleveref}}{}

\title{%
    \name{}:
    \mbox{Towards Comprehensive and Reliable}
    \mbox{Evaluation of Federated Prompt Learning}}
\author{%
    Dongping Liao\thanks{Equal contribution.} \\
    State Key Lab of IoTSC, \\
    CIS Dept, University of Macau. \\
    \texttt{yb97428@um.edu.mo} \\
    \And
    Xitong Gao\samethanks[1] \\
    Shenzhen Institutes of Advanced Technology, \\
    Chinese Academy of Sciences. \\
    Shenzhen University of Advanced Technology. \\
    \texttt{xt.gao@siat.ac.cn} \\
    \And
    Yabo Xu \\
    DataStory Information Technology Co., Ltd. \\
    \texttt{arber@datastory.com.cn} \\
    \And
    Chengzhong Xu \\
    State Key Lab of IoTSC, \\
    CIS Dept,
    University of Macau. \\
    \texttt{czxu@um.edu.mo}
}
\date{}

\begin{document}

\maketitle
\begin{abstract}
    The increasing emphasis on privacy and data security
    has driven the adoption of federated learning,
    a decentralized approach
    to train machine learning models
    without sharing raw data.
    Prompt learning,
    which fine-tunes prompt embeddings
    of pretrained models,
    offers significant advantages
    in federated settings
    by reducing computational costs and communication overheads
    while leveraging the strong performance
    and generalization capabilities
    of vision-language models such as CLIP\@.
    This paper addresses the intersection
    of federated learning and prompt learning,
    particularly for vision-language models.
    In this work,
    we introduce a comprehensive framework,
    named \name{},
    to evaluate federated prompt learning algorithms.
    \name{} assesses the performance
    of \numalgorithms{} state-of-the-art
    federated prompt learning methods
    across \numprotocols{} federated learning protocols
    and \numdatasets{} open datasets,
    considering \numscenarios{} distinct evaluation scenarios.
    Our findings demonstrate that prompt learning
    maintains strong generalization performance
    in both in-distribution and out-of-distribution settings
    with minimal resource consumption.
    This work highlights the effectiveness
    of federated prompt learning
    in environments
    characterized by data scarcity, unseen classes,
    and cross-domain distributional shifts.
    \iftoggle{final}{%
        We open-source the code
        for all implemented algorithms in \name{}
        to facilitate further research in this domain.%
        \footnote{Available at \sourcelink{}.}
    }{%
        The code for all implemented algorithms in \name{}
        is attached in supplemental material
        and will be publicly available
        to facilitate further research in this domain.
    }
\end{abstract}

\section{Introduction}\label{sec:intro}

User awareness of privacy and data security
and recent legislation
such as the General Data Protection Regulation (GDPR)
\cite{voigt2017eu,wolters2017security,politou2018forgetting}
have made it more difficult for companies
to collect and store user data.
This has led to the recent rise of federated learning
\cite{mcmahan2017communication,li2020federated}
as a promising approach
to training machine learning models
without the need to share the data itself.
Despite the potential benefits of federated learning,
it introduces large computational and communication overheads
due to the need to synchronize models training
and data heterogeneity across devices.

Leveraging large-scale text-image aligned data,
pretrained vision-language models
such as CLIP~\cite{radford2021clip}
and ALIGN~\cite{jia2021align}
show strong zero-shot image classification performance,
and have also proven to be highly effective foundational models
for various downstream tasks.
Instead of training new models,
\emph{prompt learning}~\cite{
    zhou2022coop, zhou2022cocoop, lu2022proda,
    derakhshani2023bpl, yao2023kgcoop, li2024fedotp}
fine-tunes the prompt embeddings
of these pretrained vision-language models.
This technique has demonstrated strong performance,
requiring only one or two examples per class
in the in-distribution setting,
and also excels in domain generalization.

In the context of federated learning,
which is often applied on edge devices
with constrained computational and communication capabilities,
the advantages of prompt learning are:
(1) it incurs lower computational costs
compared to training a model from scratch;
(2) it significantly
reduces communication overheads and memory requirements,
as the trainable prompt embeddings
are much smaller than the model's weights;
(3) properly utilizing the pretrained model's knowledge,
can gain strong in-distribution
and out-of-distribution generalization performance.
However,
in contrast to the centralized setting
of existing prompt learning algorithms,
we envision that federated learning
brings new perspectives specific to the paradigm
as follows:
\begin{itemize}
    \item \textbf{%
    How effective are global and personalized prompt models?}
    Inter-device data heterogeneity
    can naturally arise,
    potentially leading to slower convergence
    and degraded generalization performance of prompts.
    It remains to be seen
    how prompt learning algorithms
    can train shared global prompts
    that effectively adapt to data heterogeneity.
    Conversely,
    personalized federated learning
    trains client-specific models
    to address conflicting distributional shifts across devices
    that arise due to data heterogeneity.
    Considering the strong generalization performance
    of prompt learning,
    it is intriguing to explore
    how it may learn effective personalized prompts.

    \item \textbf{%
	Impact of various data distribution shifts
    on federated prompt learning.}
    Prompt learning algorithms
    show robust generalization capabilities
    across diverse distributional shift scenarios
    \cite{zhou2022cocoop,lu2022proda,khattak2023promptsrc}.
    This characteristic is particularly beneficial
    within federated environments,
    where devices often operate under constraints
    of limited data and computational resources.
    Our objective is to assess the effectiveness
    of federated prompt learning algorithms
    in scenarios characterized
    by the following:
    \textbf{data scarcity}
    (few-shot learning),
    \textbf{unseen classes}
    (test data containing classes not present during training),
    and \textbf{cross-domain} distributional shifts
    (test data comprising the same classes
     as training data but from different domains).

    \item \textbf{%
    Cost-effectiveness of federated prompt learning.}
    Prompt learning in federated settings
    introduces hyperparameters,
    such as the number of prompts and prompt length,
    reflecting the trade-off
    between model performance and communication cost.
    In addition,
    different baselines
    may exhibit varying convergence rates
    and have different per round
    computational and communication costs.
    Our goal is to conduct sensitivity analyses
    of various algorithms
    to examine how different setups
    may influence their respective trade-off relationship
    and algorithm choices.
\end{itemize}

Beyond providing insights to these questions,
we also aim to provide a comprehensive and extensible framework
of the federated prompt learning baselines,
and to provide a set of evaluation metrics
to assess the performance of these algorithms.
To quantify the effectiveness
of federated prompt learning algorithms
and to provide a standardized codebase
for rapid, reproducible, and reliable evaluations
in federated prompt learning,
we make the following contributions:
\begin{itemize}
    \item
    We designed and implemented \name{},
    a unified, modular and open-source codebase
    with unified training and evaluation procedures and interface,
    comprising a suite of vision-language models, datasets,
    faithful implementations of algorithms,
    and evaluation metrics.
    For ease of use,
    it has swappable modules
    for prompt learning algorithms
    and federated learning strategies,
    and zero-code configuration support
    for Hugging Face \cite{datasets} datasets and models.

    \item
    We systematically evaluated the performance
    of federated prompt learning algorithms
    under various challenging scenarios,
    including global and personalized learning
    under data heterogeneity,
    few-shot, novel-class,
    and cross-domain distributional shifts.
    Moreover,
    we explored the trade-off relationship
    between model performance and communication cost
    under different prompt learning configurations.

    \item
    Finally,
    \name{} provides a standardized comprehensive evaluation suite
    and we carried out the extensive experiments
    for the \numalgorithms{} state-of-the-art (SOTA)
    federated prompt learning baseline algorithms
    under \numprotocols{} federated learning protocols,
    with \numscenarios{} metric-reporting scenarios
    on \numdatasets{} open datasets.
    The \name{} codebase
    is fully open-source and publicly available
    for our community.
\end{itemize}

\section{Related Work and Problem Formulation}\label{sec:background}


\subsection{Related Work}\label{sec:background:related}

\textbf{%
Prompt tuning of vision-language pretrained models}
The recent advent of vision-language models (VLMs),
such as CLIP~\cite{radford2021clip}
and ALIGN~\cite{jia2021align},
which learn to align text and image pairs
in a shared embedding space using contrastive learning,
marks a significant milestone
in vision-language understanding,
showing remarkable zero-shot performance
on various downstream tasks.
Leveraging the pretrained VLMs,
CoOp~\cite{zhou2022coop}
introduces the optimization
of learnable text context embeddings
to adapt pretrained VLMs
to improve downstream performance.
CoCoOp~\cite{zhou2022cocoop}
extends CoOp by further training a Meta-Net
to predict suitable prompt embeddings
from image features.
To prevent prompt tuning
from converging to a single point,
Prompt Learning with Optimal Transport
(PLOT) \cite{chen2022plot}
formulates prompt tuning as optimal transport
between the cosine distances of visual features
and the prompt features.
Prompt Distribution Learning
(ProDA) \cite{lu2022proda}
optimizes text prompts
by learning to model Gaussian distributions
over the prompt embeddings,
and encourages semantic orthogonality
among the prompt embedding vectors
to enhance the generalization capability.
On a similar note,
Prompt-aligned gradient (ProGrad)
\cite{zhu2023prograd}
aligns the prompt gradients
that are in conflicting directions
with the zero-shot predictions,
while self-regulating prompts
(SRC) \cite{khattak2023promptsrc}
propose to condition prompted features
to be consistent with the CLIP features
with self-consistency regularization,
and knowledge-guided contextual optimization
(KgCoOp) \cite{yao2023kgcoop}
regularizes the prompt embeddings
to be within the proximity
of hand-crafted prompts.
The above three methods
aim to preserve the model's capability
on unseen classes.
Bayesian Prompt Learning (BPL)
\cite{derakhshani2023bpl}
proposes to model the prompt embeddings
as Gaussian distributions
conditioned on the image features,
and optimize the prompt embeddings
with variational inference.

\textbf{%
Federated prompt learning for vision-language model adaptation}
Federated learning \cite{mcmahan2017communication}
is an emerging paradigm
for decentralized model training
without sharing raw data,
alleviating legal and privacy concerns.
Early endeavor
in fine-tuning pretrained vision-language models
in a federated setting
has been explored in \cite{lu2023fedclip}
to address the challenges
of inter-client data heterogeneity
and improve generalization performance
under cross-domain scenarios.
Doing so, however,
incurs a heavy communication cost
of full model weights.
Instead,
prompt tuning algorithms in the federated setting
can significantly reduce communication overheads
by only transmitting the prompt embeddings,
or a small network that predicts the prompt embeddings.
PromptFL \cite{guo2023promptfl}
extends CoOp \cite{zhou2022coop}
to the federated setting
for the server aggregation of prompt embeddings
using FedAvg \cite{mcmahan2017communication}.
To mitigate feature and label shifts,
FedOTP \cite{li2024fedotp}
introduces a strategy
where a shared global prompt
is learned to extract consensus information,
along with a local prompt for each client
to capture client-specific knowledge.
Subsequently,
it applies an optimal transport-based alignment
to regularize both the global and local prompts,
balancing global consensus and local personalization.
FedTPG\cite{qiu2023text}
leverages a cross-attention module to generate
prompts conditioned on task-related text input.

\textbf{Evaluating federated learning}
Several evaluation frameworks or benchmarks
have been established for federated learning algorithms,
each offering unique perspectives.
For instance,
LEAF \cite{caldas2018leaf}
and TFF \cite{bonawitz2019tff}
tailor to heterogeneous datasets,
whereas FedML \cite{he2020fedml},
Flower \cite{beutel2020flower},
and FedScale \cite{lai2022fedscale}
emphasize diverse system resources.
Diverging from traditional tasks
such as image classification
and next word prediction,
FedNLP \cite{lin2022fednlp}
explores federated learning
with challenging natural language processing applications,
while FS-G \cite{wang2022fsg}
directs its attention to graph learning.
pFL-Bench \cite{chen2022pfl}
offers a comprehensive evaluation
of personalization in federated learning,
while FL-bench \cite{tan2023flbench}
further focuses on domain generalization.
Moreover,
FLAIR \cite{song2022flair}
offers a curated large-scale dataset
with fine-grained annotated labels
and evaluated common FL baselines using it.
Profit \cite{collins2023profit}
is related to \name{}
as it explores prompt learning
for personalization federated learning.
However,
they considered the prompt learning
of large language models
rather than vision-language foundational models
as considered in this paper.
To summarize,
the majority of the above
mostly considers general federated learning algorithms,
which cannot offer superior generalization performance
as prompt learning algorithms
under data scarcity and distribution shifts.
While it is crucial
to test known hypotheses agreed upon by the community
with standardized evaluations,
\name{} presents new and unique perspectives
in anticipation of future research questions and challenges
pertaining to federated prompt learning.

\subsection{Problem Formulation}\label{sec:background:problem}

\newcommand{\params}{\bm{\theta}}
\newcommand{\CiData}{\mathbb{D}_i}
\newcommand{\NumClient}{|\mathbb{C}|}
\newcommand{\RealSet}{\mathbb{R}}
\newcommand{\CiDataRatio}{\rho_i}

\textbf{Federated learning.}
We follow the widely-used federated averaging (FedAvg)
\cite{mcmahan2017communication}
baseline as an example
to introduce the federated training problem setup.
Specifically,
it optimizes the global objective function \(F(\params)\)
on local private data \(\CiData\)
with a total of \(\NumClient\) participating clients:
\begin{equation}
    \min_{\params \in \RealSet^d} F(\params)
    = {\sum}_{i=1}^{\NumClient} \CiDataRatio F_i(\params),
\end{equation}
where
\(
F_i(\params) = \frac{1}{|\CiData|}\sum_{\xi \in \CiData}f_i(\params)
\)
is the local learning objective of client \(i\),
and
\(\sum_{i=1}^{\NumClient}\CiDataRatio=1\).
In each communication round,
the server first transmits the updated global models
to a selected group of clients.
The selected clients
perform the local training
on private data and then transmit the trained model
to the global server.
FedAvg~\cite{mcmahan2017communication}
aggregates the models by a weighted-averaging scheme
based on the normalized number of data on each client.
Without losing generality,
we adapt several existing prompt learning
methods~\cite{zhou2022coop,zhou2022cocoop,
chen2022plot,lu2022proda,khattak2023promptsrc,
zhu2023prograd,yao2023kgcoop}
to federated training with FedAvg~\cite{mcmahan2017communication}.

\textbf{Federated prompt learning.}
Following the success
of vision-language pretrained models,
recently the FL community
has began to switch from train-from-scratch paradigm
to adapting these models
to diverse downstream tasks under federation.
Prompt learning,
a prevalent technique in NLP~\cite{gao2020making,jiang2020can,lester2021power}
has been investigated under vision applications~\cite{jia2022visual}.
Taking CLIP \cite{radford2021clip} for example,
prompt learning aims to adapt its frozen pretrained models,
consisting of an image encoder \(E_\textrm{image}\)
and a text encoder \(E_\textrm{text}\)
for downstream tasks.
In this paper,
we seek to evaluate \emph{continuous} prompt learning,
which optimizes a set of soft prompt vectors \(\{v_1, v_2, \ldots v_m\}\)
parameterized by a few learnable weights \(\theta\).
To obtain the text features of a class,
we can concatenate the learned prompt vectors
and the embedding features of a class name
to construct the class-specific embeddings \(t_j\),
followed by feeding it to a text pretrained encoder \(E_{text}\).
For classification tasks,
the prediction probability
of an input image \(x\) for class \(j\)
can be calculated by comparing the similarity
between encoded image features and text features
of each class as follows:
\begin{equation}
    p_{\params} \parens{y = j \mid x} =
        \frac{
            \exp \parens{
                \similarity \parens{
                    E_{\textrm{image}}(x),
                    E_{\textrm{text}}\parens{t_j}
                } / \tau
            }
        }{
            \sum_i^n \exp \parens{
                \similarity \parens{
                    E_{\textrm{image}}(x),
                    E_{\textrm{text}}\parens{t_i}
                } / \tau
            }
        },
\end{equation}
where \(\similarity\parens{\cdot, \cdot}\) represents a metric function
such as dot product or cosine similarity,
and \(\tau\) is a scaling factor
to control the temperature of Softmax operation.
To optimize the prompt vectors,
federated prompt learning
enables the communication-efficient FL training
by synchronizing the prompt parameters across clients
under the coordination of a global server.
Keeping true to conventional federated training
which transmits trained parameters,
federated prompt learning
communicates the prompt-related parameters
between the server and client for iterative optimization.

\section{Framework Design}\label{sec:framework}

\begin{figure*}[t]
    \centering
    \includegraphics[
        width=\linewidth, trim=20pt 30pt 10pt 45pt, clip
    ]{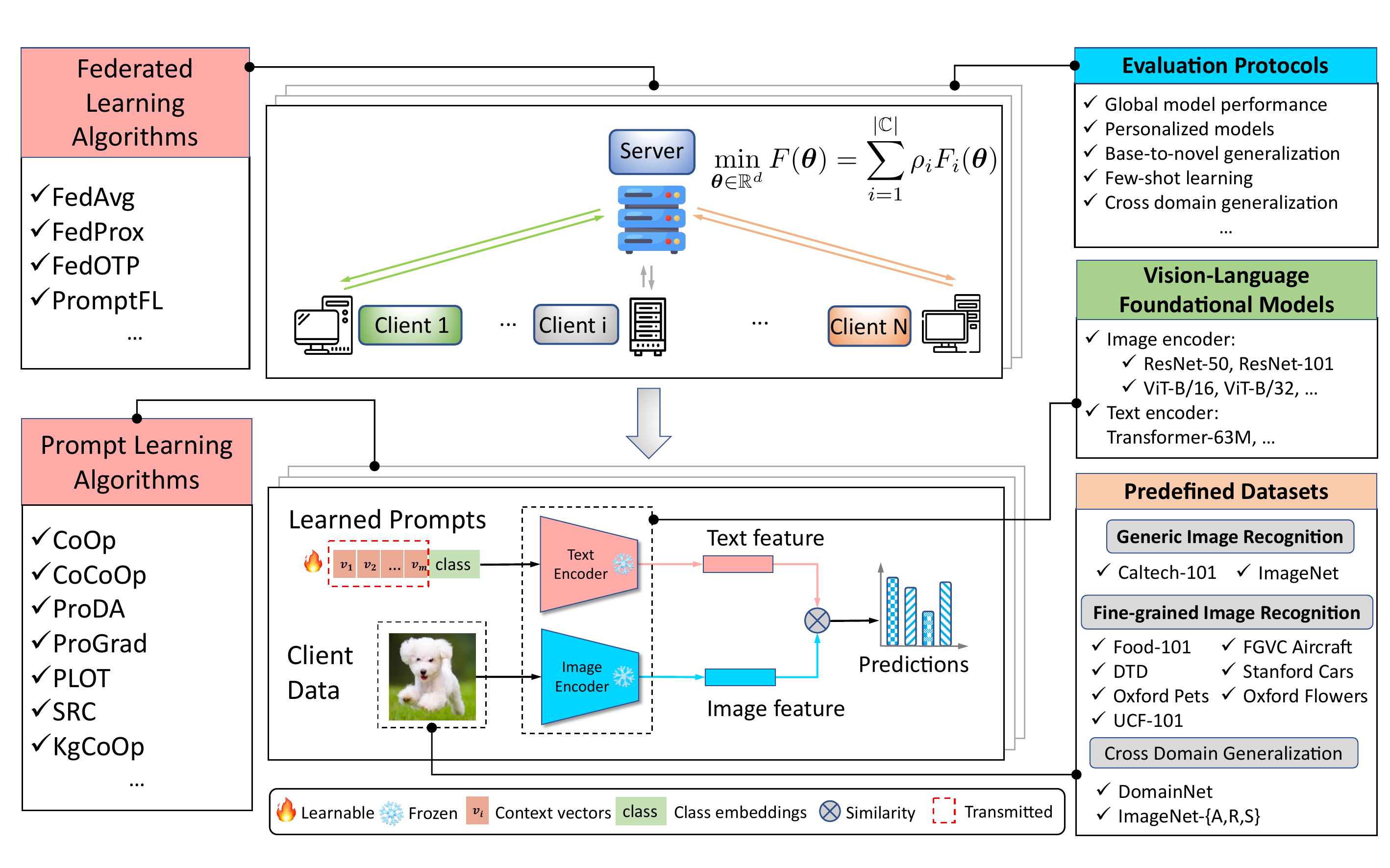}
    \caption{%
        Overview of the \name{} framework.
    }\label{fig:overview}
    \cvspace{-10pt}
\end{figure*}

\Cref{fig:overview} provides an overview
of the \name{} framework design.
This section details the models
and \numdatasets{} datasets used,
\numalgorithms{} algorithmic baselines
for federated prompt learning,
and \numprotocols{} evaluation protocols
with various metrics
under \numscenarios{} evaluation scenarios.

\subsection{Models and Datasets}\label{sec:framework:models_datasets}

\textbf{Models}
All algorithms
adopts pretrained CLIP \cite{radford2021clip} models
as the base models for prompt learning.
CLIP models
are vision-language models
that consist of an image encoder \(E_\textrm{image}\)
and a text encoder \(E_\textrm{text}\),
both of which map the input data
into a shared feature space.
The architecture of the image encoder
can either be a ResNet-50 \cite{he2016resnet}
or Vision Transformer (ViT) \cite{dosovitskiy2021vit},
whereas the text encoder
adopts a transformer \cite{vaswani2017transformer} architecture.
In the main experiments,
we used the CLIP model based on ResNet-50.

\textbf{Datasets}
For the main evaluations,
we used 8 publicly available datasets  
with annotated labels
for global shared models \textaccglobal{}
and personalized models \textaccpersonal{}:
\textbf{Caltech}-101 \cite{fei2004caltech},
\textbf{DTD} \cite{cimpoi2014dtd},
FGVC-\textbf{Aircraft} \cite{maji2013aricraft},
\textbf{Food}-101 \cite{bossard2014food},
Stanford\textbf{Cars} \cite{krause2013cars},
Oxford-\textbf{Flowers}-102 \cite{nilsback2008flowers},
Oxford-\textbf{Pets} \cite{parkhi2012pets},
and \textbf{UCF}-101 \cite{soomro2012ucf101}.
For novel class and few-shot scenarios,
we use the first 4 datasets
consisting of generic and fine-grained
image recognition tasks.
For brevity,
we use the bolded part of the dataset names
as the shorthand in the results.
For cross-domain metrics,
\Cref{app:results:cross}
further examines the robustness
of federated prompt learning algorithms
on distribution shifts.
For which
ImageNet \cite{deng2009imagenet}
is used as the source domain,
and the target domains
include variants of ImageNet-derived datasets:
\textbf{ImageNet-S}ketch \cite{wang2019sketch}
(images transformed to sketches)
\textbf{ImageNet-A}dversarial \cite{hendrycks2021nae}
(natural adversarial examples in the wild),
and \textbf{ImageNet-R}endition \cite{hendrycks2021robustness}
(images with diverse styles).

\subsection{Algorithmic Baselines}\label{sec:framework:baselines}

We include PromptFL \cite{guo2023promptfl}
and FedOTP \cite{li2024fedotp},
two recent SOTA federated prompt learning methods
in our comparison.
To further enrich baseline comparisons,
we adapt all centralized prompt tuning algorithms
mentioned in \Cref{sec:background:related}
to the federated setting
to form each federated variant,
prefixed with ``\fed{}'',
namely \fed{CoCoOp}, \fed{PLOT},
\fed{ProDA}, \fed{ProGrad}, \fed{SRC},
and \fed{KgCoOp}.
While we provide other aggregation strategies
in our codebase,
for a fair comparison in this paper,
all variants use federated averaging (FedAvg)
\cite{mcmahan2017communication}
to aggregate prompt-related parameters on the server.
Finally,
we include ZS-CLIP,
\ie{}, the pretrained CLIP \cite{radford2021clip}
as a zero-shot baseline
(\ie{}, with hand-crafted prompts)
to provide a reference point.

\subsection{Evaluation Protocols and Metrics}\label{sec:framework:eval}

\textbf{Prompt learning protocols}
We set the prompt context token length as 4.
We try to align the number of prompts for methods evaluated.
Specifically,
we use a single set of prompt as default,
with the exceptions that
FedOTP naturally uses two set of prompts
for aligning the global and local representation
and \fed{ProDA} adopts two set of prompts
for prompt distribution learning.
For all evaluations,
we fix the class token position in the \emph{end}
without token position augmentation.
We only learn the prompt
for the text input unless otherwise stated.
For specific hyperparameter settings
related to prompt learning,
please refer to \Cref{app:setup:hyper}.

\textbf{Federated learning protocols}
We evaluate the federated prompt learning algorithms
under \numprotocols{} federated learning protocols
as follows:
\begin{itemize}
    \item
    \textbf{Standard learning}
    We simulate 10 clients with full participation
    to train a global model to evaluate on a shared test set.
    We use the standard SGD optimizer
    with initial learning rate 0.002,
    momentum 0.9 and a cosine learning rate decay scheduler
    to guarantee the convergence of each method.
    We set the batch size as 16,
    global communication rounds to 50,
    and the local training epoch to 1.
    For other detailed experimental hyperparameter settings
    under this protocol,
    please refer to \Cref{app:setup:hyper}.

    \item \textbf{Partial participation}
    For the client sub-sampling protocol,
    we follow the standard learning above
    and increase the number of clients
    to 100 with a \( 10\% \) participation ratio
    to investigate the scalability
    of federated prompt learning algorithms.

    \item \textbf{Personalized learning}
    This protocol seeks to evaluate the adaptability
    of federated prompt learning methods
    to local data distribution for personalization purpose.
    Following~\cite{xu2022personalized},
    we split a test dataset sharing similar distribution
    of the training dataset for evaluation.
    The personalized learning protocol
    uses similar settings as standard learning,
    except that the performance is evaluated
    on personalized test sets
    instead of a shared test set.

    \item
    \textbf{Centralized learning}
    For reference purposes,
    we also evaluate the performance
    of the prompt learning algorithms in centralized settings,
    where all clients' data is aggregated
    and trained in a centralized manner.
    This setting mimics the traditional centralized prompt learning.
\end{itemize}
Finally,
each experiment was conducted with \numtrials{} runs
with different random seeds.
We report the best test accuracy on test set
with mean value and standard deviation.

\textbf{Evaluation scenarios}
We use the following \numscenarios{} scenarios
where each reports evaluation metrics
to assess the performance of federated prompt learning algorithms
with \emph{different data splits and evaluation settings}.
Here,
each scenario adopts the standard learning protocol
unless otherwise stated.
\begin{itemize}
    \item
    \textbf{Global shared prompt learning}
    (reports: \textaccglobal{})
    We split the training, validation and testing
    samples following~\cite{zhou2022cocoop}.
    We apply a Dirichlet data partition scheme
    with the concentration parameter be set as \(0.1\)
    that produces \noniid{} data sets for local clients
    following~\cite{wang2019federated}.
    This metric is evaluated
    under the standard learning
    and partial participation protocols.

    \item
    \textbf{Personalized prompt learning}
    (reports: \textaccpersonal{})
    The accuracy is reported by a weighted averaging
    of local clients' accuracy
    based on the number of data each client possesses.
    This is more robust than a direct averaging
    of local clients' accuracy as it reduces the impact
    of the potential fluctuating accuracy of some clients
    that hold very few training and testing samples.
    This metric is evaluated
    under the personalized learning protocol.

    \item \textbf{Base-to-novel class generalization}
    (reports: \textaccbase{}, \textaccnovel{})
    As prompt learning algorithms
    has the ability to generalize to unseen classes,
    we evaluate the FL variants
    on novel class generalization
    following the protocol below.
    First,
    we split all classes into two equal sets,
    where only the first half (containing base classes)
    is used for training,
    following the standard learning protocol
    with \noniid{} data heterogeneity.
    After training,
    we evaluate the model
    on the test sets of both halves
    (respectively containing base and novel classes).
    We report the accuracy on both sets,
    denoted as \textaccbase{} and \textaccnovel{}
    respectively,
    along with the harmonic mean
    of the two accuracies,
    namely, \(
        \accharmonic \triangleq
        2 / (\accbase^{-1} + \accnovel^{-1})
    \).

    \item
    \textbf{Few-shot learning}
    (reports: \textaccfewshot{K})
    For few-shot generalization evaluation,
    where each client
    has only a small training shot \( K \) per class,
    we apply \iid{} sampling
    to draw \( K \) samples per class
    for each client from the training set.

    \item \textbf{Cross-domain generalization}
    (reports: \textaccxd{x}{y},
    where ``x'' and ``y''
    denote the source and target domain datasets,
    respectively.)
    To evaluate the robustness
    of federated prompt learning algorithms
    on distribution shifts,
    we evaluate the performance on the cross-domain datasets
    with shared classes but distinctive domain distributions.
    Our training setting
    aligns with the standard learning protocol
    with \noniid{} data heterogeneity,
    using a source dataset (\eg{}, ImageNet) for training,
    and a target dataset (\eg{}, ImageNet-\{A,R,S\}) for testing.
    Due to page limits,
    we report the cross-domain accuracies
    in \Cref{app:results:cross}
    and full results in our online codebase.

    \item \textbf{Cost-performance trade-offs}
    (reports:
    global accuracy \textaccglobal{},
    communication cost \textcommcost{})
    Finally,
    certain prompt learning-specific hyperparameters
    (\eg{}, the number of prompts and prompt length)
    can influence the trade-off relationship
    between model performance and system resources,
    it is thus important
    to investigate how these hyperparameters
    affect the performance of federated prompt learning algorithms,
    in terms of both converged model performance,
    computational and communication costs.
    We conducted sensitivity analyses of various algorithms
    on these hyperparameters
    to explore both how they influence
    the trade-off relationship,
    and provide insights for sweet-spot configurations.
    We report the relevant results
    in \Cref{app:results:cost}
    and provide the full results in our codebase.
\end{itemize}

Beyond these accuracy metrics,
we are also concerned with the
stability of evaluated algorithms.
Motivated by the ``Ranking Scores"
from the Out-of-Distribution Generalization
literature~\cite{ye2022ood, huang2022two},
we introduce a \textbf{superiority indicator},
which counts the total number of datasets
on which a method's performance
surpasses the baseline (PromptFL).
The corresponding results are shown by
columns with "\textbf{\#}" header in each table.

\section{Experimental Results}\label{sec:experiments}

We conduct extensive experiments
under various FL scenarios and evaluation metrics.
Below we present some important results
on global and personalized performance,
base-to-novel class generalization
and few-shot learning capabilities.
These metrics are mostly concerned
in federated learning and prompt learning research.
\begin{table*}[ht]
    \centering
    \caption{%
        \textbf{%
        Comparison of shared global model accuracy
        \textaccglobal{} (\%)
        of federated prompt learning methods.}
        We report the mean \( \pm{} \) standard deviation
        over \numtrials{} runs.
        The best and second-best results
        for each dataset are highlighted
        in \textbf{bold} and \underline{underlined},
        respectively.
        The ``\textbf{\#}'' column
        indicates on how many datasets
        the method achieves performance
        exceeding \textbf{PromptFL}.
    }\label{tab:global}
    \adjustbox{max width=\linewidth}{%
    \begin{tabular}{l|rrrrrrrr|r|c}
        \toprule
        \multicolumn{1}{c|}{Global \textaccglobal{}}
            & \theader{Caltech} & \theader{DTD}
            & \theader{Aircraft} & \theader{Food}
            & \theader{Cars} & \theader{Flowers}
            & \theader{Pets} & \theader{UCF}
            & \multicolumn{1}{|c|}{\textbf{Avg.}} & \theader{\#} \\
        \midrule
        \textbf{ZS-CLIP}
            & 86.0 & 41.7
            & 16.6 & 77.9
            & 55.5 & 65.3
            & 85.7 & 61.5
            & 61.3 & - \\
        \midrule
        \textbf{PromptFL}
            & \tpm{91.5}{0.5} & \tpm{57.6}{1.3}
            & \tpm{22.8}{0.4} & \tpm{79.2}{0.1}
            & \tpm{62.0}{0.4} & \tbpm{84.0}{1.7}
            & \tbpm{89.4}{0.5} & \tpm{70.1}{0.8}
            & \underline{69.6} & - \\
        \textbf{FedOTP}
            & \tupm{91.8}{0.1} & \tpm{58.0}{0.8}
            & \tpm{21.9}{0.4} & \tpm{78.7}{0.1}
            & \tbpm{62.8}{0.2} & \tpm{83.3}{0.6}
            & \tpm{89.1}{0.1} & \tpm{69.4}{0.6}
            & 69.4 & 3 \\
        \midrule
        \textbf{\fed{CoCoOp}}
            & \tpm{91.7}{0.3} & \tpm{54.7}{1.0}
            & \tpm{17.9}{4.5} & \tpm{79.3}{0.1}
            & \tpm{60.7}{0.5} & \tpm{76.4}{0.7}
            & \tpm{89.1}{0.1} & \tpm{68.0}{0.9}
            & 67.2 & 2 \\
        \textbf{\fed{PLOT}}
            & \tpm{91.6}{0.3} & \tbpm{58.3}{1.4}
            & \tpm{21.7}{0.5} & \tpm{78.3}{0.2}
            & \tpm{60.7}{0.8} & \tpm{83.4}{0.6}
            & \tpm{88.9}{0.5} & \tpm{69.7}{0.4}
            & 69.1 & 2 \\
        \textbf{\fed{ProDA}}
            & \tpm{91.6}{0.3} & \tpm{57.2}{1.1}
            & \tbpm{23.1}{0.7} & \tpm{79.1}{0.2}
            & \tpm{62.3}{0.5} & \tbpm{84.0}{0.8}
            & \tupm{89.3}{0.4} & \tpm{70.2}{1.1}
            & \underline{69.6} & \textbf{5} \\
        \textbf{\fed{ProGrad}}
            & \tpm{90.7}{0.2} & \tpm{57.1}{1.0}
            & \tpm{21.7}{0.3} & \tbpm{79.5}{0.1}
            & \tpm{60.5}{0.7} & \tpm{83.4}{0.4}
            & \tpm{89.1}{0.2} & \tupm{70.3}{0.3}
            & 69.1 & 2 \\
        \textbf{\fed{PromptSRC}}
            & \tbpm{92.0}{0.8} & \tpm{57.8}{0.3}
            & \tpm{21.2}{0.4} & \tpm{78.6}{0.4}
            & \tupm{62.4}{0.2} & \tpm{83.6}{0.0}
            & \tpm{89.2}{0.7} & \tupm{70.3}{1.0}
            & 69.4 & \underline{4} \\
        \textbf{\fed{KgCoOp}}
            & \tupm{91.8}{0.2} & \tupm{58.2}{0.8}
            & \tupm{23.0}{0.1} & \tupm{79.4}{0.2}
            & \tpm{61.7}{0.7} & \tupm{83.9}{0.5}
            & \tbpm{89.4}{0.2} & \tbpm{70.4}{0.7}
            & \textbf{69.7} & \textbf{5} \\
        \bottomrule
    \end{tabular}}
\end{table*}

\begin{table*}[ht]
    \centering
    \caption{%
        \textbf{%
        Comparison of personal model accuracy
        \textaccpersonal{} (\%)
        of federated prompt learning methods
        on various datasets.}
    }\label{tab:personal}
    \adjustbox{max width=\linewidth}{%
    \begin{tabular}{l|rrrrrrrr|r|c}
        \toprule
        \multicolumn{1}{c|}{Personal \textaccpersonal{}}
            & \theader{Caltech} & \theader{DTD}
            & \theader{Aircraft} & \theader{Food}
            & \theader{Cars} & \theader{Flowers}
            & \theader{Pets} & \theader{UCF}
            & \multicolumn{1}{|c|}{\textbf{Avg.}} & \theader{\#} \\
        \midrule
        \textbf{ZS-CLIP}
            & 86.0 & 41.7
            & 16.6 & 77.9
            & 55.5 & 65.3
            & 85.7 & 61.5
            & 61.3 & - \\
        \midrule
        \textbf{PromptFL}
            & \tpm{91.5}{0.4} & \tpm{69.5}{4.1}
            & \tpm{33.8}{0.1} & \tbpm{82.1}{0.4}
            & \tpm{67.7}{0.6} & \tbpm{89.7}{0.2}
            & \tbpm{89.9}{0.6} & \tpm{77.5}{1.5}
            & 75.2 & 0 \\
        \textbf{FedOTP}
            & \tbpm{91.9}{0.4} & \tbpm{73.8}{1.4}
            & \tbpm{36.1}{0.5} & \tupm{82.0}{0.7}
            & \tbpm{68.1}{1.7} & \tupm{89.6}{0.3}
            & \tupm{89.5}{1.2} & \tbpm{80.7}{1.0}
            & \textbf{76.5} & \underline{5} \\
        \midrule
        \textbf{\fed{CoCoOp}}
            & \tupm{91.8}{0.4} & \tpm{70.3}{3.0}
            & \tpm{34.0}{1.7} & \tpm{81.8}{0.6}
            & \tpm{67.4}{0.7} & \tpm{86.4}{1.9}
            & \tupm{89.5}{0.9} & \tpm{77.1}{0.9}
            & 74.8 & 3 \\
        \textbf{\fed{PLOT}}
            & \tpm{91.7}{0.4} & \tupm{71.3}{3.1}
            & \tpm{34.0}{0.8} & \tpm{81.4}{1.2}
            & \tupm{67.9}{1.1} & \tpm{89.3}{0.5}
            & \tpm{88.6}{0.3} & \tupm{79.6}{1.8}
            & \underline{75.5} & \underline{5} \\
        \textbf{\fed{ProDA}}
            & \tpm{91.7}{0.9} & \tpm{69.7}{2.6}
            & \tupm{34.7}{0.6} & \tbpm{82.1}{1.3}
            & \tpm{67.8}{1.3} & \tpm{89.3}{0.6}
            & \tbpm{89.9}{0.5} & \tpm{77.9}{1.5}
            & 75.4 & \textbf{6} \\
        \textbf{\fed{ProGrad}}
            & \tpm{91.7}{0.6} & \tpm{69.2}{0.3}
            & \tpm{32.9}{0.7} & \tpm{81.6}{0.8}
            & \tpm{67.0}{1.0} & \tpm{88.8}{0.9}
            & \tupm{89.5}{0.7} & \tpm{77.0}{1.5}
            & 74.7 & 1 \\
        \textbf{\fed{PromptSRC}}
            & \tpm{91.7}{0.4} & \tpm{69.3}{1.3}
            & \tpm{32.4}{2.3} & \tpm{81.8}{1.1}
            & \tpm{67.6}{1.1} & \tpm{89.2}{1.7}
            & \tpm{89.3}{1.7} & \tpm{78.2}{1.1}
            & 74.9 & 2 \\
        \textbf{\fed{KgCoOp}}
            & \tpm{91.6}{0.3} & \tpm{68.6}{2.7}
            & \tpm{31.3}{0.5} & \tpm{81.4}{0.7}
            & \tpm{67.1}{1.4} & \tpm{88.9}{0.9}
            & \tbpm{89.9}{0.3} & \tpm{76.9}{0.9}
            & 74.5 & 1 \\
        \bottomrule
    \end{tabular}}
    \cvspace{-10pt}
\end{table*}

\begin{table*}[t]
\centering\caption{%
    \textbf{Comparison of base and novel class accuracy \((\%)\)
    of federated prompt learning methods.}
    Statistics are aggregated with 10 experimental runs
    under different random splits of base and novel classes.
    In each independent run,
    we align the base and novel classes
    across different evaluated methods
    for fair comparison.
    Note that ``\textaccbase{}'', ``\textaccnovel''
    and ``\textaccharmonic{}'' rows
    correspond to base \textaccbase{}
    and novel class \textaccnovel{} accuracies,
    and their harmonic mean
    \( \accharmonic \triangleq 2 / (\accbase^{-1} + \accnovel^{-1}) \),
    respectively.
    Results are reported
    in similar style as \Cref{tab:global}
    for \textaccharmonic{} values.
}\label{tab:base2novel_random}
\cvspace{-0.5em}
\setlength{\tabcolsep}{2.5pt}
\adjustbox{max width=\linewidth}{%
\begin{tabular}{l|ccc|ccc|ccc|ccc|ccc|c}
    \multicolumn{1}{c}{}
        & \multicolumn{3}{c}{\textbf{Caltech}}
        & \multicolumn{3}{c}{\textbf{Aircraft}}
        & \multicolumn{3}{c}{\textbf{Cars}}
        & \multicolumn{3}{c}{\textbf{Flowers}}
        & \multicolumn{3}{c}{\textbf{Avg.}}
        & \\
    \toprule
    Metric
        & \textaccbase & \textaccnovel & \textaccharmonic
        & \textaccbase & \textaccnovel & \textaccharmonic
        & \textaccbase & \textaccnovel & \textaccharmonic
        & \textaccbase & \textaccnovel & \textaccharmonic
        & \textaccbase & \textaccnovel & \textaccharmonic
        & \theader{\#} \\
    \midrule
    \textbf{ZS-CLIP}
        & 88.2  & 92.6  & 90.3
        & 19.6  & 24.7  & 21.8
        & 59.5  & 68.1  & 63.5
        & 77.2  & 71.0  & 73.9
        & 61.1  & 64.1  & 62.4
        & -\\
    \midrule
    \textbf{PromptFL}
        & \tgpm{92.8}{0.8}  & \tgpm{92.7}{0.7}  & \tpm{92.6}{0.2}
        & \tgpm{20.9}{0.4}  & \tgpm{24.7}{0.5}  & \tpm{22.6}{0.2}
        & \tgpm{63.0}{0.7}  & \tgpm{67.6}{0.7}  & \tpm{65.2}{0.0}
        & \tgpm{79.9}{2.9}  & \tgpm{69.3}{0.7}  & \tpm{74.2}{1.0}
        & \txtg{64.1}   & \txtg{63.8}  & 63.8   & -\\
    \textbf{FedOTP}
        & \tgpm{93.1}{0.2}  & \tgpm{93.7}{0.4}  & \tupm{93.4}{0.1}
        & \tgpm{21.0}{0.8}  & \tgpm{23.7}{1.0}  & \tpm{22.2}{0.8}
        & \tgpm{62.1}{0.1}  & \tgpm{66.0}{0.7}  & \tpm{64.0}{0.4}
        & \tgpm{81.0}{0.6}  & \tgpm{68.7}{1.9}  & \tpm{74.3}{1.2}
        & \txtg{64.3}   & \txtg{63.0}  & 63.5   & 2\\
    \midrule
    \textbf{\fed{CoCoOp}}
        & \tgpm{92.7}{0.8}  & \tgpm{93.6}{0.6}  & \tpm{93.1}{0.2}
        & \tgpm{18.0}{2.1}  & \tgpm{17.1}{2.4}  & \tpm{17.2}{2.2}
        & \tgpm{62.8}{0.6}  & \tgpm{66.7}{0.3}  & \tpm{64.7}{0.2}
        & \tgpm{79.4}{1.4}  & \tgpm{70.7}{1.8}  & \tupm{74.8}{0.6}
        & \txtg{63.2}   & \txtg{62.0}  & 62.4   & 2\\
    \textbf{\fed{PLOT}}
        & \tgpm{93.4}{0.6}  & \tgpm{93.5}{1.0}  & \tbpm{93.5}{0.8}
        & \tgpm{19.0}{0.8}  & \tgpm{23.4}{0.4}  & \tpm{21.0}{0.4}
        & \tgpm{62.4}{0.5}  & \tgpm{65.4}{1.4}  & \tpm{63.8}{0.7}
        & \tgpm{78.7}{3.0}  & \tgpm{68.3}{1.3}  & \tpm{73.1}{0.6}
        & \txtg{63.4}   & \txtg{62.6}  & 62.8   & 1\\
    \textbf{\fed{ProDA}}
        & \tgpm{93.0}{0.4}  & \tgpm{92.7}{1.1}  & \tpm{92.8}{0.4}
        & \tgpm{22.0}{0.5}  & \tgpm{25.1}{1.1}   & \tupm{23.4}{0.6}
        & \tgpm{63.3}{0.7}  & \tgpm{67.7}{0.4}   & \tpm{65.4}{0.2}
        & \tgpm{77.9}{0.7}  & \tgpm{69.6}{0.5}   & \tpm{73.5}{0.1}
        & \txtg{64.0}   & \txtg{63.8}  & 63.8   & \underline{3}\\
    \textbf{\fed{ProGrad}}
        & \tgpm{93.2}{0.4}  & \tgpm{93.0}{0.4}  & \tpm{93.1}{0.4}
        & \tgpm{21.6}{0.5}  & \tgpm{25.2}{1.6}   & \tpm{23.3}{0.4}
        & \tgpm{64.2}{0.6}  & \tgpm{67.9}{0.5}   & \tbpm{66.0}{0.3}
        & \tgpm{80.3}{1.1}  & \tgpm{70.6}{0.5}   & \tbpm{75.2}{0.2}
        & \txtg{64.7}   & \txtg{64.2}  & \textbf{64.3}   & \textbf{4}\\
    \textbf{\fed{SRC}}
        & \tgpm{90.2}{0.2}  & \tgpm{93.0}{0.1}  & \tpm{91.6}{0.1}
        & \tgpm{21.7}{1.0}  & \tgpm{24.9}{1.0}  & \tpm{23.1}{0.8}
        & \tgpm{61.2}{0.2}  & \tgpm{67.1}{0.3}  & \tpm{64.0}{0.1}
        & \tgpm{78.9}{1.1}  & \tgpm{70.2}{0.8}  & \tpm{74.3}{0.7}
        & \txtg{62.3}   & \txtg{63.8}  & 62.8   & \underline{3}\\
    \textbf{\fed{KgCoOp}}
        & \tgpm{93.5}{0.4}  & \tgpm{93.4}{1.0}  & \tupm{93.4}{0.5}
        & \tgpm{22.3}{0.7}  & \tgpm{25.1}{0.5}  & \tbpm{23.6}{0.5}
        & \tgpm{63.7}{0.3}  & \tgpm{67.5}{0.3}  & \tupm{65.6}{0.0}
        & \tgpm{79.6}{2.7}  & \tgpm{68.9}{1.2}  & \tpm{73.9}{0.5}
        & \txtg{64.8}   & \txtg{63.7}  & \underline{64.1}   & \underline{3} \\
    \bottomrule
\end{tabular}}
    \cvspace{-10pt}
\end{table*}

\begin{table}[t]
    \centering
    \caption{%
        \textbf{Comparison of few-shot training accuracy
        \textaccfewshot{1} (\%)
        of federated prompt learning methods
        on various datasets.}
    }\label{tab:fewshot}
    \adjustbox{max width=\linewidth}{%
    \setlength{\tabcolsep}{2.5pt}
    \begin{tabular}{l|rrrr|r|c}
        \toprule
        \multicolumn{1}{c|}{Few-shot \textaccfewshot{1}}
            & \theader{Caltech} & \theader{Aircraft}
            & \theader{Cars} & \theader{Flowers}
            & \multicolumn{1}{|c|}{\textbf{Avg.}} & \theader{\#} \\
        \midrule
        \textbf{ZS-CLIP}
            & 86.0 & 16.6
            & 55.5 & 65.3
            & 55.9 & - \\
        \midrule
        \textbf{PromptFL}
            & \tpm{88.7}{0.3} & \tpm{17.3}{1.0}
            & \tpm{55.7}{0.2} & \tpm{65.3}{1.2}
            & \underline{57.2} & - \\
        \textbf{FedOTP}
            & \tbpm{89.8}{0.4} & \tpm{17.8}{1.2}
            & \tbpm{56.8}{0.2} & \tupm{65.6}{0.8}
            & \textbf{57.5} & \textbf{4} \\
        \midrule
        \textbf{\fed{CoCoOp}}
            & \tpm{87.6}{0.6} & \tpm{17.6}{0.5}
            & \tpm{55.4}{0.2} & \tpm{64.6}{1.3}
            & 56.3 & 1 \\
        \textbf{\fed{PLOT}}
            & \tpm{87.5}{0.7} & \tpm{17.6}{0.6}
            & \tpm{55.5}{0.1} & \tbpm{65.7}{1.3}
            & 56.6 & 2 \\
        \textbf{\fed{ProDA}}
            & \tpm{89.0}{0.2} & \tpm{17.3}{0.6}
            & \tpm{56.1}{0.7} & \tbpm{65.7}{1.3}
            & 57.0 & 3 \\
        \textbf{\fed{ProGrad}}
            & \tupm{89.4}{0.6} & \tupm{18.4}{0.2}
            & \tpm{56.2}{0.3} & \tpm{63.5}{0.6}
            & 56.9 & 3 \\
        \textbf{\fed{SRC}}
            & \tpm{89.2}{0.3} & \tbpm{18.9}{0.3}
            & \tupm{56.4}{0.5} & \tpm{65.4}{0.0}
            & \textbf{57.5} & \textbf{4} \\
        \textbf{\fed{KgCoOp}}
            & \tpm{88.4}{0.4} & \tpm{18.3}{0.8}
            & \tupm{56.4}{0.1} & \tpm{62.9}{0.6}
            & 56.5 & 2 \\
        \bottomrule
    \end{tabular}}
\end{table}
\iftoggle{arxiv}{%
\begin{table}[t]
\centering\caption{%
    \textbf{Comparison of few-shot accuracies
    \textaccfewshot{k} (\%)
    of federated prompt learning methods
    on Tiny-ImageNet.}
    Number of shots \( k \in \braces{1,2,4,8,16} \).
    Results are reported
    in similar style as \Cref{tab:global}.
    ZS-CLIP has an accuracy of 34.1\%.
}\label{tab:fewshot_sweep}
\cvspace{-10pt}
\adjustbox{max width=\linewidth}{%
\begin{tabular}{l|ccccc|c|c}
    \toprule
    Few-shot \textaccfewshot{\cdot}
        & \textbf{1} & \textbf{2}
        & \textbf{4} & \textbf{8}
        & \textbf{16}
        & \textbf{Avg.} & \textbf{\#} \\
    \midrule
    \textbf{PromptFL}
        & \tpm{39.9}{0.4} & \tpm{42.1}{0.2}
        & \tpm{44.0}{0.2} & \tpm{45.8}{0.8}
        & \tpm{47.0}{0.2}
        & 43.7 & - \\
    \textbf{FedOTP}
        & \tupm{41.0}{0.2} & \tbpm{43.9}{0.1}
        & \tbpm{46.4}{0.2} & \tbpm{48.2}{0.2}
        & \tbpm{49.4}{0.1}
        & \textbf{45.8} & \textbf{5} \\
    \midrule
    \textbf{\fed{CoCoOp }}
        & \tpm{38.2}{0.4} & \tpm{41.9}{0.2}
        & \tpm{43.3}{0.3} & \tpm{45.4}{0.3}
        & \tpm{45.3}{0.2}
        & 42.8 & 0 \\
    \textbf{\fed{PLOT   }}
        & \tpm{39.5}{0.2} & \tpm{41.7}{0.1}
        & \tpm{43.2}{0.4} & \tpm{45.3}{0.4}
        & \tpm{46.3}{0.1}
        & 43.2 & 0 \\
    \textbf{\fed{ProDA  }}
        & \tpm{40.3}{0.1} & \tpm{42.0}{0.3}
        & \tpm{44.0}{0.2} & \tpm{45.6}{0.3}
        & \tupm{47.2}{0.3}
        & 43.8 & 2 \\
    \textbf{\fed{ProGrad}}
        & \tpm{35.6}{0.8} & \tpm{36.7}{0.4}
        & \tpm{38.6}{0.2} & \tpm{38.8}{0.4}
        & \tpm{39.8}{0.4}
        & 37.9 & 0 \\
    \textbf{\fed{SRC    }}
        & \tbpm{41.6}{0.4} & \tupm{42.6}{0.5}
        & \tupm{44.4}{0.5} & \tpm{45.0}{0.2}
        & \tpm{46.1}{0.2}
        & \underline{43.9} & \underline{3} \\
    \textbf{\fed{KgCoOp }}
        & \tpm{39.6}{0.4} & \tpm{42.0}{0.6}
        & \tpm{43.6}{0.9} & \tupm{46.3}{0.2}
        & \tpm{46.5}{0.3}
        & 43.6 & 1 \\
    \bottomrule
\end{tabular}}

\end{table}
\begin{figure}[t]
    \centering\includegraphics[
        width=0.5\linewidth, trim=30pt 20pt 25pt 25pt
    ]{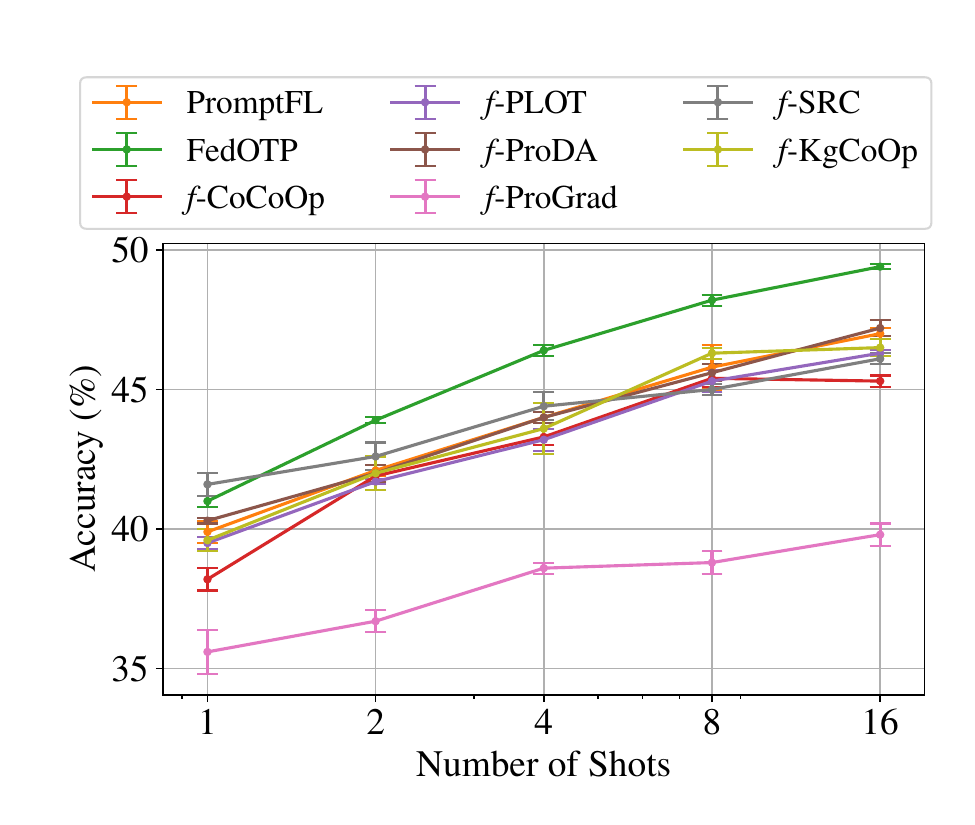}
    \captionof{figure}{%
        \textbf{Few-shot accuracies
        of federated prompt learning methods
        on Tiny-ImageNet.}
    }\label{fig:fewshot_sweep}
\end{figure}
}{%
\begin{table*}[t]
    \begin{minipage}{0.66\linewidth}
        
    \end{minipage}\hfill%
    \begin{minipage}{0.30\linewidth}
        \centering\includegraphics[
            width=\linewidth, trim=30pt 40pt 25pt 45pt
        ]{shots}
        \captionof{figure}{%
            \textbf{Few-shot accuracies
            of federated prompt learning methods
            on Tiny-ImageNet.}
        }\label{fig:fewshot_sweep}
    \end{minipage}
    \cvspace{-5pt}
\end{table*}
}

\begin{table}[ht]
\centering\caption{%
    Comparison of accuracy (\%)
    on various datasets
    under \textbf{client subsampling}
    with 10\% of the total 100 clients.
}\label{tab:client_subsample_0.1}
\adjustbox{max width=\linewidth}{%
\begin{tabular}{l|cccc|c|c}
    \toprule
        & \textbf{Caltech} & \textbf{Aircraft}
        & \textbf{Cars} & \textbf{Flowers}
        & \textbf{Avg.} & \textbf{\#} \\
    \midrule
    \textbf{ZS-CLIP}
            & 86.0 & 16.6
            & 55.5 & 65.3
            & 55.9 & - \\
    \midrule
    PromptFL
        & \tpm {90.3}{0.2}      & \tpm {20.9}{0.7}
        & \tpm {60.9}{0.5}      & \tpm {75.9}{1.3}
        & 62.0    & - \\
    FedOTP
        & \tupm {91.1}{0.6}      & \tpm {21.2}{0.8}
        & \tpm {59.1}{0.1}      & \tupm {76.2}{0.9}
        & 61.9    & \underline{3} \\
    \midrule
    \fed{CoCoOp}
        & \tpm {90.4}{0.5}      & \tpm {17.5}{1.6}
        & \tpm {59.7}{0.5}      & \tpm {73.9}{0.7}
        & 60.4    & 1 \\
    \fed{PLOT}
        & \tpm {90.6}{0.2}      & \tpm {20.5}{0.6}
        & \tpm {59.1}{0.7}      & \tpm {74.8}{1.5}
        & 61.3    & 1 \\
    \fed{ProDA}
        & \tpm {90.8}{0.5}      & \tpm {21.7}{0.4}
        & \tupm {61.0}{0.6}      & \tpm {75.0}{0.7}
        & \underline{62.1}    & \underline{3} \\
    \fed{ProGrad}
        & \tpm {90.7}{0.1}      & \tbpm {22.2}{0.6}
        & \tpm {60.3}{0.5}      & \tpm {74.6}{0.2}
        & 61.9    & 2 \\
    \fed{SRC}
        & \tpm {90.6}{2.0}      & \tpm {21.9}{1.4}
        & \tbpm {61.5}{0.2}      & \tupm {76.2}{0.8}
        & \textbf{62.6}    & \textbf{4} \\
    \fed{KgCoOp}
        & \tbpm {91.2}{0.1}      & \tupm {22.0}{0.6}
        & \tpm {60.4}{0.3}      & \tbpm {76.7}{1.1}
        & \textbf{62.6}    & \underline{3} \\
    \bottomrule
\end{tabular}}
\end{table}

\textbf{Global shared prompt learning}
In~\Cref{tab:global},
we illustrate a comprehensive evaluation
of the performance of global models
across \numglobaldatasets{} datasets.
We summarize some key insights below:
\begin{itemize}
    \item First,
    after aligning the experimental settings,
    the performance gaps
    between the baseline PromptFL
    and various prompt learning methods
    are less significant
    compared with the results reported
    in existing federated prompt learning literatures
    \cite{guo2023promptfl,li2024fedotp}.
    Indeed,
    PromptFL \cite{guo2023promptfl},
    a simple combination of CoOp and FedAvg,
    serves as a very strong baseline
    for other federated prompt learning methods,
    occasionally achieving the best performance
    on fine-grained image recognition datasets
    such as OxfordPets and Flowers
    over \eval{\numalgorithms{} - 1} competitors.
    We advocate acknowledging its simplicity and merits
    and including it as a reference baseline
    in all federated prompt learning works.

    \item Second,
    in most cases,
    the \fed{CoCoOp} produces inferior results
    compared with PromptFL baseline.
    We hypotheses this is caused by its adoption
    of an image feature aggregation module,
    which is susceptible to the data heterogeneity
    raised by \noniid{} data partitions.
    As a result,
    its aggregated features
    may deviate from real class semantics
    if the biased local training
    is not properly counteracted.
    This underscores the potential risks
    of a direct porting of centralized prompt learning
    methods to federated learning regime.

    \item Third,
    from the superiority indicator,
    we can observe the regularization-based
    federated prompt learning methods,
    such as \fed{SRC} and \fed{KgCoOp}
    generally produce discernible improvements
    for FL generic performance.
    It demonstrates such regularization
    can yield a favorable effect
    to reduce the local client drift~\cite{karimireddy2020scaffold}.
    Specifically,
    this regularization enforces all participating clients
    share a common objective that encourages
    the learning of domain knowledge
    by introduced a prescribed text prompt.
    This highlights the similar intuitions
    behind the regularization-based federated learning
    such as FedProto~\cite{tan2022fedproto}
    and prompt learning methods
    exemplified by PromptSRC~\cite{khattak2023promptsrc}.
\end{itemize}

\textbf{Personalized prompt learning}
\Cref{tab:personal} presents the personalized
performance comparison.
Interestingly,
we find FedOTP~\cite{li2024fedotp}
generally outperforms other methods,
emphasizing the potential of distribution
alignment, for example, with Optimal Transport (OT)
to adapt to personalized data distribution.
The results also indicate
FedOTP consistently outperforms
a baseline method \fed{PLOT},
which also applying OT to align
representations across modalities,
demonstrating advantage
of \emph{imbalanced} Optimal Transport
over \fed{PLOT}
for personalized FL scenarios.
This implies in addition to modality
gap between vision and text representations,
the distribution gap under federated prompt learning
introduces additional challenges to be addressed.
Besides,
indicated by the superiority metric,
we observe the improvements of regularization-based prompt
learning methods under personalized data
are less prominent compared with
the results in~\Cref{tab:global}.
This could be an intrinsic dilemma
that achieving improvements
on global and personalized performance
could compromise each other
and requires further exploration.

\textbf{Base-to-novel generalization}
We illustrated the results
on base \textaccbase{}
and novel \textaccnovel{} class accuracies,
along with their harmonic mean \textaccharmonic{}
in~\Cref{tab:base2novel_random}.
Similar to global learning,
\Cref{tab:base2novel_random}
indicates regularization with generic prompt knowledge
\cite{khattak2023promptsrc,zhu2023prograd,yao2023kgcoop}
prevents the client model
from over-fitting the local data distribution,
and it contributes to achieving the best trade-offs
between base class fitting and novel class generalization.
Different from the global scenario,
the na{\"\i}ve PromptFL is less effective
in handling this challenging scenario
according to the ``\textbf{\#}'' competition metric.

\textbf{Few-shot generalization}
The results on few-shot generalization
in~\Cref{tab:fewshot} stress the difficulties
of existing methods in terms of better generalization
over the PromptFL baseline with a few training samples.
Intriguingly,
all three methods that achieve superior performance,
either use multiple prompt sets
(FedOTP~\cite{li2024fedotp} and~\fed{ProDA}~\cite{lu2022proda})
or sample multiple prompts
(\fed{SRC}~\cite{khattak2023promptsrc})
during the inference stage.
These results hint the effectiveness
of ensembling the knowledge of multiple prompts,
which can be useful to reduce the \emph{sample selection bias}
raised by limited training samples
\cite{xu2022alleviating,yang2021bridging}.
Focusing on varying number of shots,
\Cref{tab:fewshot_sweep}
shows the few-shot training accuracies
of federated prompt learning methods
on Tiny-ImageNet.
Notably,
we sweep the number of shots
with \( k \in \braces{1, 2, 4, 8, 16} \).
\Cref{fig:fewshot_sweep}
also visualizes the few-shot training accuracies
of the methods.
In short,
FedOTP shows the strongest ability
to learn from limited training samples,
owing to its global and local cooperative prompt design
and distribution alignment optimization.

\textbf{Client Sub-sampling}
In cross-device FL system,
the partial participation of massive clients
could have a detrimental forgetting effect
on the global model
due to their temporarily joining and quitting FL training.
In~\Cref{tab:client_subsample_0.1},
we explored this effect
by scaling up to 100 simulated clients
with a \( 10\% \) participating ratio.
We observe that under this scenario,
the regularization-based federated prompt learning methods
usually gain advantages over the baselines,
this could be attributed to the regularization loss
which introduces a common optimization objective among clients,
alleviating the catastrophic forgetting.

\begin{figure}[t]
    \centering
    \iftoggle{arxiv}{%
        \includegraphics[width=0.8\linewidth]{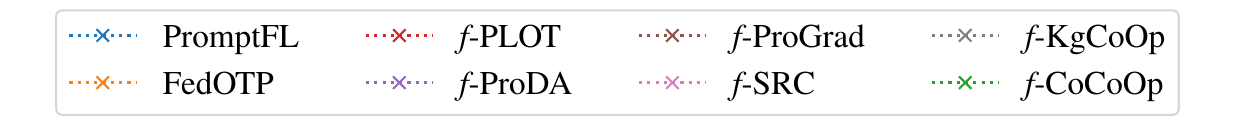}
    }{%
        \includegraphics[width=\linewidth, trim=20pt 0 20pt 0]{legend}
    }
    \begin{subfigure}{0.5\linewidth}
        \centering
        \iftoggle{arxiv}{%
            \includegraphics[
                width=\linewidth, trim=0 20pt 0 0,
            ]{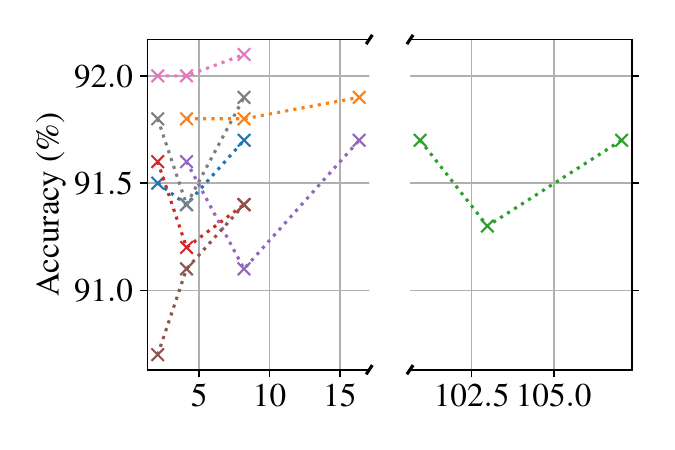}%
        }{%
            \includegraphics[
                width=\linewidth, trim=20pt 20pt 10pt 20pt,
            ]{prompts}%
        }
        \caption{%
            Number of prompts: \( \{ 1, 2, 4 \} \).
        }\label{fig:cost:prompts}
    \end{subfigure}%
    \begin{subfigure}{0.5\linewidth}
        \centering
        \iftoggle{arxiv}{%
            \includegraphics[
                width=\linewidth, trim=0 20pt 0 0,
            ]{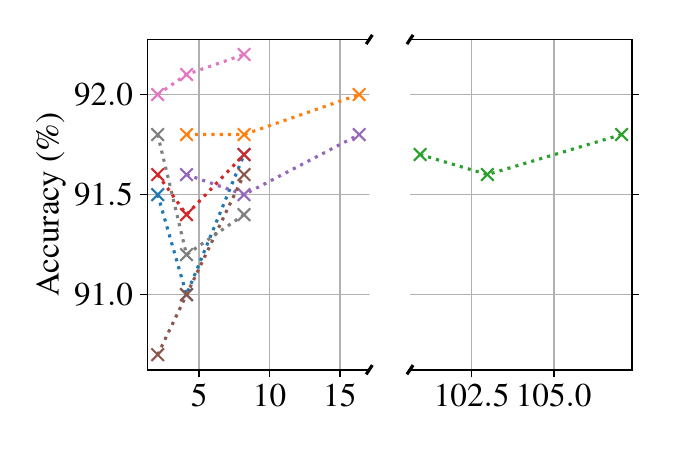}%
        }{%
            \includegraphics[
                width=\linewidth, trim=20pt 20pt 10pt 20pt,
            ]{tokens}%
        }
        \caption{%
            Number of tokens: \( \{ 4, 8, 16 \} \).
        }\label{fig:cost:tokens}
    \end{subfigure}%
    \caption{%
        \textbf{Trade-offs between accuracy (\%)
        and the number of communicated parameters (in millions)
        on Caltech-101.}
    }\label{fig:cost}
    \cvspace{-15pt}
\end{figure}

\textbf{Cost-performance Trade-offs}
\Cref{fig:cost:prompts,fig:cost:tokens}
present the communication and performance trade-offs
by changing the number of prompts
and prompt token lengths respectively.
Please refer to \Cref{tab:cost:prompts,tab:cost:tokens}
in \Cref{app:results:cost}
for tabulated results.
First,
we note that a direct scaling of the learnable parameters
does not necessarily deliver positive improvements.
For example,
\fed{CoCoOp} employs a meta-net
to aggregate the conditional image information,
which drastically increases the number of communications.
However,
this does not translates
to accuracy boost over the simple baseline
in most experiments.
Besides,
by comparing the accuracies increments of a single methods
with different number of prompt or prompt length,
we can conclude that methods
with distribution alignment (\fed{FedOTP})
or diversity regularization (\fed{ProDA})
usually bring in stable improvements
when scaling up the prompt parameters.
Finally,
both approaches for tweaking the prompt parameters
yield similar improvements.
Indeed,
we do not observe clear dominance of them
over the other.

\textbf{More results}
Due to space limitation,
we present additional results
in \Cref{app:results},
specifically:
centralized training (\Cref{app:results:central})
and cross-domain generalization (\Cref{app:results:cross}).
Moreover,
in \Cref{app:results:cost}
we show that \name{}
fully support the cost-performance trade-off exploration
between accuracies and training costs,
and provide results on the trade-off relationship.

\begin{figure}[t]
\cvspace{-1em}
\begin{minipage}{\linewidth}
    \centering
    \definecolor{FrameColor}{RGB}{4, 165, 229}
    \begin{tcolorbox}[
        enhanced, clip upper,
        noparskip,
        width=0.96\linewidth,
        colback=FrameColor!15,
        colframe=FrameColor!50,
        arc=1mm,
        auto outer arc,
        left=0.5em,
        right=0.5em,
        title={\bfseries\color{black}Key Takeaways},
    ]
\begin{itemize}
    \item \textbf{Global shared prompt learning}
    PromptFL, combining CoOp and FedAvg,
    serves as a simple yet effective baseline
    and often achieves the best performance
    on fine-grained image recognition datasets
    like Oxford-Pets and Flowers.

    \item \textbf{Personalized Prompt Learning}
    FedOTP generally outperforms other
    personalized prompt learning methods,
    highlighting the efficacy of distribution alignment
    in adapting to personalized data.

    \item \textbf{Base-to-novel Generalization}
    Regularization prevents overfitting
    and balances base and novel class metrics,
    without it (PromptFL) is less effective.

    \item \textbf{Few-shot Generalization}
    Methods that use multiple prompts
    (\eg{}, FedOTP, \fed{ProDA}, \fed{SRC})
    perform best in few-shot scenarios,
    indicating ensembling helps reduce sample selection bias.

    \item \textbf{Client Sub-sampling}
    The regularization-based methods
    are favorable owing to
    alleviating the catastrophic forgetting.

\end{itemize}
\end{tcolorbox}
\cvspace{-1em}
\end{minipage}
\end{figure}

\section{Conclusion}\label{sec:conclusion}

This paper presents \name{},
the first comprehensive evaluation
for federated prompt learning algorithms.
Through extensive experiments
on various datasets and evaluation scenarios,
we demonstrate the effectiveness of prompt learning
in federated settings,
particularly in challenging scenarios
characterized by data scarcity, unseen classes,
and cross-domain distributional shifts.
\name{}
provides a standardized and extensible open-source codebase,
complete with evaluation metrics and various open datasets,
facilitating further research in this promising area.
\name{} serves as a valuable tool
for researchers and practitioners
to explore the trade-offs
between model performance and system resources
in federated prompt learning,
paving the way for the development
of more efficient and effective algorithms.

{%
    \small
    \bibliographystyle{ieeenat_fullname}
    \bibliography{references}
}
\appendix
\section{Datasets}\label{app:setup:data}

We evaluated federated prompt learning methods
on \numdatasets{} datasets,
including generic image recognition,
fine-grained image recognition
and domain generalization tasks.
These datasets are prevalent
in the FL and prompt learning literature,
covering a wide range of scales, domains and partitions.
For the detailed experimental setup steps
of these datasets,
please refer to the instructions from our project page.
Below,
we briefly summarize these datasets.

\subsection{Generic Image Recognition}

For generic image recognition,
we considered the following three datasets.
\begin{itemize}
    \item The \textbf{Caltech-101} \cite{fei2004caltech}
    contains 9,146 images
    divided into 101 distinct object categories,
    along with a background category.
    Each object category has between 40 to 800 images,
    with most categories having around 50 images.
    The images are of variable sizes
    and depict objects in various poses
    and viewpoints against different backgrounds.

    \item The \textbf{ImageNet} dataset \cite{deng2009imagenet}
    we used in our experiments
    contains over 1.2 million labeled images
    spread across 1,000 categories.

    \item The \textbf{Tiny-ImageNet} dataset
    is a downsized variant of the ImageNet dataset.
    It consists of 200 distinct classes,
    each containing 500 training images,
    50 validation images,
    and 50 test images,
    summing up to 100,000 images in total.
\end{itemize}

\subsection{Fine-grained Image Recognition}

For fine-grained image recognition,
we evaluated 7 datasets:
\begin{itemize}
    \item The \textbf{Describable Textures Dataset (DTD)} \cite{cimpoi2014dtd}
    is a collection of images specifically designed
    for studying texture recognition task.
    It contains 5,640 images categorized into 47 classes,
    each representing a distinct texture
    described by human-centric attributes
    like ``bumpy,'' ``striped,''
    or ``polka-dotted.''
    Each class includes 120 images
    sourced from diverse environments,
    ensuring variability in appearance.
    The DTD is commonly used to develop and evaluate algorithms
    for recognizing and classifying textures
    based on their describable properties.

    \item The \textbf{FGVC Aircraft} dataset \cite{maji2013aricraft}
    is a specialized collection of images
    aimed at fine-grained visual classification of aircraft models.
    It includes 10,000 images of aircraft,
    encompassing 100 different aircraft model variants.
    Each image is annotated with detailed information
    such as aircraft type, variant, and manufacturer.
    This dataset is used to develop and evaluate algorithms
    that can distinguish between visually similar aircraft models,
    making it valuable for applications
    requiring precise classification
    within a narrowly defined category.

    \item The \textbf{Food-101} dataset \cite{bossard2014food}
    is a collection of images
    designed for food recognition tasks in computer vision.
    It contains 101 categories of food,
    with each category represented by 1,000 images,
    totaling 101,000 images.
    The images are split into a training set of 75,750 images
    and a test set of 25,250 images.
    This dataset is used to develop and test algorithms
    that can accurately identify different types of food,
    making it useful for applications
    in dietary tracking, culinary automation,
    and food-related research.

    \item The \textbf{Oxford Pets} dataset \cite{parkhi2012pets}
    is a collection of images
    for the fine-grained classification and segmentation
    of pet breeds.
    It includes 37 categories,
    encompassing different breeds of cats and dogs,
    with roughly 200 images per category.
    Each image is annotated with breed labels
    and additional information like pixel-level segmentation masks.
    This dataset is used to develop and evaluate algorithms
    for pet breed identification and object segmentation.

    \item The \textbf{Oxford Flowers} dataset \cite{beutel2020flower}
    is a collection of images designed
    for flower classification tasks.
    It comprises 102 flower categories,
    with each category containing between 40 and 258 images,
    totaling 8,189 images.
    The images are annotated with class labels,
    making the dataset suitable
    for automated botanical identification.

    \item \textbf{Stanford Cars} \cite{krause2013cars}
    includes 16,185 images of 196 car models,
    covering a variety of makers, models, and years.
    The dataset is divided
    into 8,144 training images and 8,041 testing images.

    \item \textbf{UCF Action Recognition} dataset \cite{soomro2012ucf101}
    is a widely used dataset for action recognition in videos.
    It consists of thousands of video clips
    collected from YouTube across 101 action categories,
    such as walking, running, and jumping.
    In our experiments,
    we use the static frames of each action for prompt learning,
    and evaluate the model performance on a disjoint test set.
\end{itemize}

\subsection{Domain Generalization}

\begin{figure}[ht]
    \centering
    \foreach \datasetname in {A,R,S}{%
        \begin{subfigure}{\linewidth}
            \centering
            \foreach \x in {1,...,4}{%
                \includegraphics[height=0.17\linewidth]{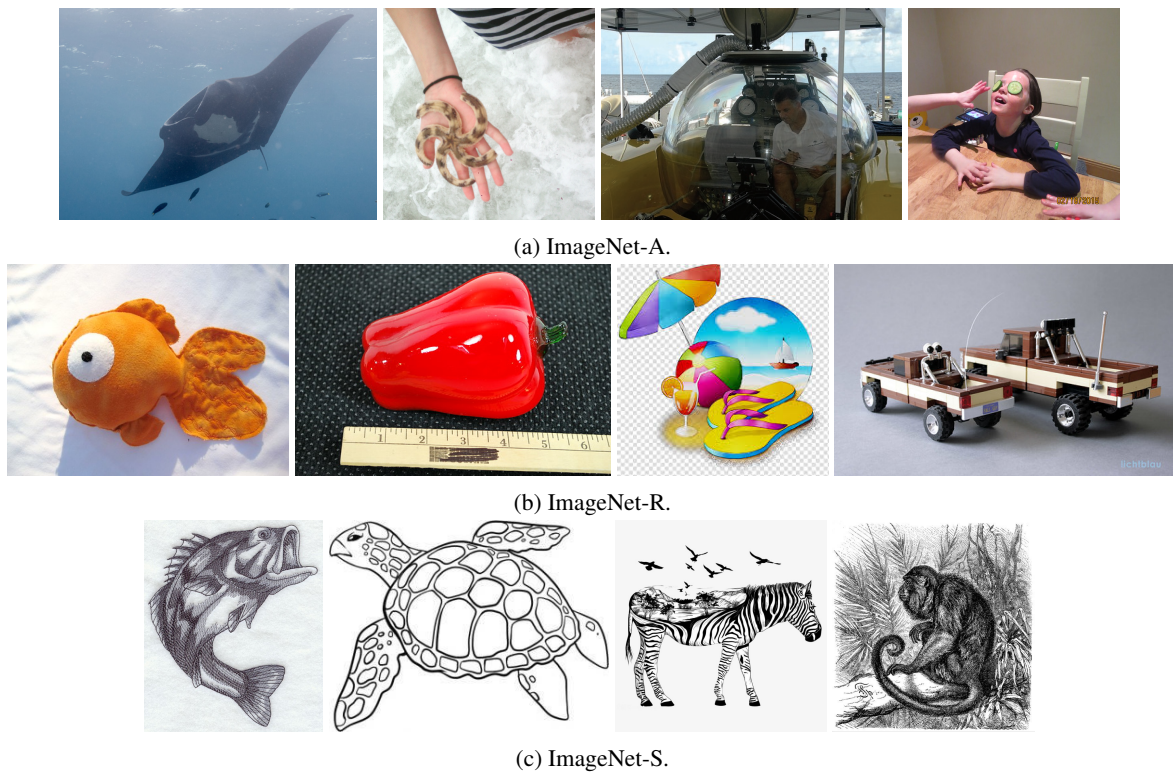}
            }
            \caption{ImageNet-\datasetname.}\label{fig:examples:\datasetname}
        \end{subfigure}
    }\caption{%
        Example images sampled from ImageNet-A, ImageNet-R, and ImageNet-S.
    }\label{fig:examples:xdomain}
\end{figure}

To evaluate the domain generalization ability
of these algorithms,
we optimize the prompt on ImageNet
and further evaluate the obtained model
on three datasets with domain shifts:
\begin{itemize}
    \item The \textbf{ImageNet-A(dvarsarial)} dataset \cite{hendrycks2021nae}
    introduces 7,500 \emph{testing} images
    for 200 ImageNet classes.
    It contains real-world, unmodified,
    and naturally occurring examples
    that can significantly degrade the performance
    of machine learning models.

    \item The \textbf{ImageNet-R(endition)} dataset
    \cite{hendrycks2021robustness}
    consists of the renditions of ImageNet images
    such as art, cartoons and graffiti.
    It contains 200 ImageNet classes,
    with each class 150 images,
    resulting in total 30,000 images.

    \item The \textbf{ImageNet-S(ketch)} dataset \cite{wang2019sketch}
    comprises 50,889 images,
    with about 50 images
    corresponding to each of the 1,000 ImageNet classes.
    These images are obtained through Google Image searches
    using the query "sketch of a \texttt{<class>}".
    The search is restricted
    to the "black and white" color scheme.
    Initially,
    100 images are queried for each class,
    followed by manual curation
    to remove irrelevant and similar images.
    In cases where fewer than 50 images remain after cleaning,
    the dataset is augmented through image flipping and rotation.
\end{itemize}
As ImageNet-A and ImageNet-R
contain only a fractional of classes
from the original ImageNet dataset,
we curated a subset of ImageNet
that contains the images
belonging to the 200 classes of training data.
For ImageNet-S dataset,
we use all 1,000 classes
of ImageNet as the training data.
Examples of these datasets
are shown in \Cref{fig:examples:xdomain}.

\section{Training Details}\label{app:setup}

\subsection{Models}\label{app:setup:models}

To align with previous works~\cite{guo2023promptfl},
we adapt the pretrained models from CLIP \cite{radford2021clip}
to federated prompt learning.
Specifically,
we use a ResNet-50 model as the image feature encoder
and a Transformer~\cite{vaswani2017attention} model
with 63 million parameters and 8 attention heads
as the textual feature encoder.
We freeze the weights of image and text encoders
and only tune the learnable prompt for \emph{text input}.

\subsection{Data Heterogeneity}\label{app:setup:heter}

\subsubsection{Feature-Shift Data Heterogeneity}

\begin{figure*}[h]
    \centering
    \includegraphics[width=\linewidth]{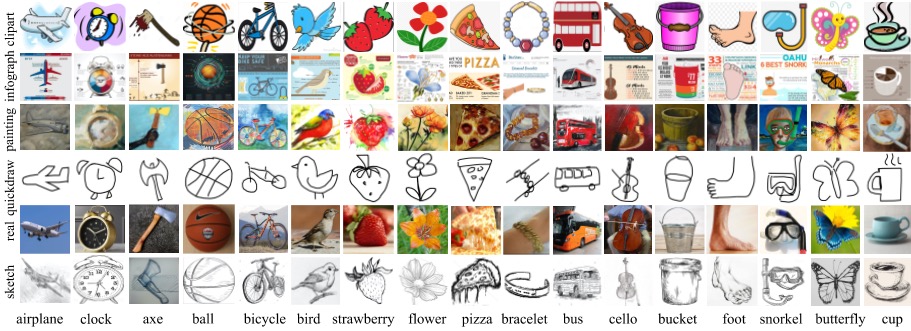}
    \caption{%
        An illustration of images
        sampled from DomainNet \cite{peng2019moment}%
        \protect\footnotemark{}.
    }\label{fig:domainnet}
\end{figure*}

In a FL system,
the participating clients
may collect data from distinct domains.
This introduce the training-time feature shift data heterogeneity
that could hinder the generalization of obtained models.
In light of this,
we evaluate the resilience
of federated prompt learning algorithms
under such data heterogeneity
in addition to label distribution data heterogeneity.
We use the \textbf{DomainNet}~\cite{peng2019moment} dataset
consisting of six domains,
each representing a distinct visual domain
such as \textbf{Clipart}, \textbf{Painting}, \textbf{Real},
\textbf{Quickdraw}, \textbf{Infograph}, and \textbf{Sketch}.
\Cref{fig:domainnet}
exemplifies the images sampled from these domains.
For each domain,
we assign the training data to two clients,
resulting in 12 clients
with each client only possessing training data
from a single domain.
\footnotetext{\url{https://ai.bu.edu/M3SDA/}.}

\subsubsection{Class-Shift Data Heterogeneity}

We adopt two \textbf{data partition strategies} for local clients,
\ie, the IID and non-IID,
according to FL scenarios or metrics we evaluated:
\begin{itemize}
    \item The \textbf{IID} dataset partition is only applied
        to evaluate the fewshot generalization performance
        of global models.
        Concretely,
        we assign each client
        equal training samples of each class
        based on the number of shots
        to evaluate the performance under limited training data.
    \item The \textbf{non-IID} dataset partition
        is applied to evaluate the performance of methods
        under label distribution data heterogeneity.
        We take the partition strategy from~\cite{wang2019federated}
        to simulate heterogeneous data partition
        based on a Dirichlet distribution
        controlled by a concentration parameter \( \alpha \).
        A smaller \( \alpha \)
        indicates more aggressive data heterogeneity,
        while an infinitely large \( \alpha \)
        degenerates to the \iid{} data partition.
        We set \( \alpha=0.1 \) by default in all experiments
        unless otherwise stated.
        To partition the original training data for multiple clients,
        we first subsample a smaller balanced dataset
        from the original training set,
        and then apply the Dirichlet data partition.
        For all experiments,
        we subsample 8 images for each class
        to construct the dataset for partition
        except for few-shot experiments mentioned above.
        Besides,
        we set 16 images for each class
        for partial participation scenarios.
\end{itemize}
Notably,
the experimental setting on evaluating DomainNet
introduces additional feature shift data heterogeneity
raised by domain-specific features on each client.
This can be considered as an extension
on the \noniid{} data partition with domain heterogeneity.

\subsection{Methods}\label{app:setup:methods}

We consider PromptFL \cite{guo2023promptfl}
and FedOTP \cite{li2024fedotp}
as two baselines of federated prompt learning.
For more comprehensive comparison,
we also include centralized methods,
such as CoCoOp, PLOT, ProDA, ProGrad, Prompt-SRC and KgCoOp.
All centralized methods
are implemented for local client training
and combined with FedAvg~\cite{mcmahan2017communication}
for global aggregation.
The evaluation helps to understand the properties
of existing prompt learning methods
under a broad range of evaluation metrics for federated training.
This also conveys insights
of the appropriate application scenarios
of each method under federation.

Below we make a brief introduction
of federated learning methods:
\begin{itemize}
    \item \textbf{PromptFL} \cite{guo2023promptfl}
        is a simple yet effective
        federated prompt learning method
        that can be viewed as a federated variant
        of CoOp \cite{zhou2022coop}
        with FedAvg~\cite{mcmahan2017communication}
        for global aggregation.
        PromptFL only communicates the shared soft prompts
        instead of a shared global model as in conventional FL.
        This drastically reduces the communication cost of FL.
    \item \textbf{FedOTP}\footnote{\url{https://github.com/HongxiaLee/FedOTP}.}
        \cite{li2024fedotp}
        extends upon the PLOT~\cite{chen2022plot}
        and designs a novel optimization scheme
        for imbalanced optimal transport.
        It also proposes
        to learn both local and global aligned representation
        for better generalization.
        The FedOTP is originally designed for personalized FL,
        to evaluate its performance on generic FL,
        we make a small alteration
        to allow all local prompt parameters
        to be communicated and updated.
\end{itemize}
In addition,
here are the original centralized prompt learning methods
that we adapt to the federated setting:
\begin{itemize}
    \item \textbf{CoCoOp}\footnote{\url{https://github.com/KaiyangZhou/CoOp}.}
    \cite{zhou2022cocoop} addresses the base and novel
    generalization dilemma of prompt learning
    by introducing conditional inference.
    It optimizes an additional meta-net
    to deliver the image features
    for complementing the domain specific information
    during inference,
    which alleviates the over-fitting
    issue of the CoOp~\cite{zhou2022coop}.

    \item \textbf{PLOT}\footnote{\url{https://github.com/CHENGY12/PLOT}.}
    \cite{chen2022plot} introduces optimal transport
    to match the image and textual features.
    This benefits the distribution alignment
    of cross-modality features
    to reduce the modality gaps.

    \item \textbf{ProDA}\footnote{\url{https://github.com/bbbdylan/proda} (unofficial).}
    \cite{lu2022proda} proposes a optimization framework
    to learn a Gaussian distribution over possible prompts
    rather than relying on a single static prompt.
    It also prompts the diversity of prompt sets
    by introducing a orthogonality regularization loss term.

    \item \textbf{ProGrad}\footnote{\url{https://github.com/BeierZhu/Prompt-align.}}
    \cite{zhu2023prograd}
    updates the prompt with aligned gradient (or non-conflicting)
    to the general knowledge which is achieved by regularizing
    gradient update with tailored prompts for domain-specific dataset.

    \item \textbf{KgCoOp}\footnote{\url{https://github.com/htyao89/KgCoOp}.}
    \cite{yao2023kgcoop}
    is a concurrent work
    that also introduces tailored prompts for each dataset
    as an anti-overfitting technique
    for guiding the prompt optimization.
    This reduces the discrepancy between the textual features
    produced by learnable prompts and the hand-crafted prompts,
    enhancing the generalization ability for unseen classes.

    \item \textbf{SRC}\footnote{\url{https://github.com/muzairkhattak/PromptSRC}.}
    \cite{khattak2023promptsrc}
    regularizes the prompt learning
    with the predictions of the frozen model,
    multiple prompts over the training trajectory
    and textual diversity from different prompt templates.
    It reduces the catastrophic forgetting
    of generalizable knowledge
    from the pretrained CLIP models.
\end{itemize}

To unify the experimental settings
for fair and faithful results,
we adapt the official public code implementation
of these centralized methods into our FL framework
with minimal alterations such as renaming the arguments.
For methods that lacked public code,
we either re-implement their algorithms
or port the unofficial code
with careful scrutinizing of the algorithmic details
to ensure alignment with original papers.
We will continuously
include more federated prompt learning methods
into \name{}\@.

\subsection{Hyperparameters}\label{app:setup:hyper}

We use the standard SGD optimizer
with initial learning rate 0.002,
momentum 0.9,
and a cosine learning rate decay scheduler
to guarantee the sufficient convergence of each method.
We set the batch size as 16,
global communication rounds 50
and the local training epoch 1.
For each run,
we use \emph{random} prompt initialization
without prompt position augmentation
when constructing the entire prompt
with prefix, class name and suffix.
The input images is resized to \( 240\times240 \)
then cropped with size \( 224\times224 \)
to match the input image size
of CLIP~\cite{radford2021clip} image encoder,
followed by random horizontal flipping
and normalization with mean and variance
calculated on the entire dataset.
For each experiments,
we conduct 3 independent runs
with different random seeds
and report the mean and standard variance
of accuracy on test set.

\section{Additional Results}\label{app:results}

\subsection{Centralized setting}\label{app:results:central}

In~\Cref{tab:central},
we report the training accuracy values
of prompt learning methods under the centralized setting.
\emph{With the initialization of the pretrained models,
there is only a slender margin between centralized and federated settings}.
We speculate the underlying reason
is that rich features from the pretrained
models significantly reduce the potential
gradient conflict among client updates.
This observation holds the promise of
closing the gap between centralized and federated training,
motivating practical and efficient algorithms
that specifically seek out better generalization
with pretrained vision-language models.
\begin{table*}[h]
\centering\caption{%
    \textbf{Comparison of training accuracy (\%)
    of prompt learning methods
    under the centralized (\ie{} non-FL) setting.}
    We report the mean \( \pm{} \) standard deviation
    over \numtrials{} runs.
}\label{tab:central}
\adjustbox{max width=\linewidth}{%
\begin{tabular}{l|rrrrrrrr}
    \toprule
    \multicolumn{1}{c|}{Centralized}
        & \theader{Caltech} & \theader{DTD}
        & \theader{Aircraft} & \theader{Food}
        & \theader{Cars} & \theader{Flowers}
        & \theader{Pets} & \theader{UCF} \\
    \midrule
    \textbf{ZS-CLIP}    & 86.0             & 41.7                 & 16.6             & 77.9                 & 55.5                  & 65.3               & 85.7             & 61.5             \\
    \midrule
    \textbf{CoOp}       & \tpm{91.5}{0.8}  & \tpm{58.1}{1.0}      & \tpm{23.5}{0.8}  & \tpm{79.3}{0.3}      & \tpm{63.0}{0.2}       & \tpm{86.4}{0.1}    & \tpm{89.3}{0.5}  & \tpm{70.7}{0.3}  \\
    \textbf{CoCoOp}     & \tpm{91.9}{0.2}  & \tpm{57.2}{1.0}      & \tpm{19.1}{1.4}  & \tpm{79.4}{0.5}      & \tpm{62.7}{0.2}       & \tpm{79.9}{1.5}    & \tpm{88.9}{0.2}  & \tpm{68.7}{1.5}  \\
    \textbf{PLOT}       & \tpm{91.7}{0.3}  & \tpm{58.8}{0.4}      & \tpm{23.4}{0.8}  & \tpm{78.3}{0.1}      & \tpm{62.4}{0.6}       & \tpm{86.1}{0.2}    & \tpm{89.6}{0.3}  & \tpm{71.0}{0.1}  \\
    \textbf{ProDA}      & \tpm{91.8}{0.3}  & \tpm{57.0}{0.8}      & \tpm{22.8}{0.2}  & \tpm{79.0}{0.2}      & \tpm{63.6}{0.6}       & \tpm{88.6}{0.7}    & \tpm{89.0}{0.2}  & \tpm{70.9}{0.5}  \\
    \textbf{ProGrad}    & \tpm{91.2}{0.3}  & \tpm{57.8}{1.0}      & \tpm{21.7}{1.3}  & \tpm{79.4}{0.1}      & \tpm{63.3}{0.1}       & \tpm{87.9}{0.3}    & \tpm{89.1}{1.2}  & \tpm{70.1}{0.9}  \\
    \textbf{PromptSRC}  & \tpm{92.2}{0.1}  & \tpm{57.9}{1.6}      & \tpm{22.7}{0.2}  & \tpm{78.9}{0.1}      & \tpm{63.5}{0.2}       & \tpm{84.2}{3.2}    & \tpm{89.4}{0.1}  & \tpm{71.5}{0.3}  \\
    \textbf{KgCoOp}     & \tpm{91.8}{0.2}  & \tpm{58.6}{0.5}      & \tpm{23.8}{0.1}  & \tpm{79.5}{0.3}      & \tpm{64.3}{0.6}       & \tpm{84.3}{2.0}    & \tpm{89.6}{0.7}  & \tpm{71.3}{0.7}  \\
    \bottomrule
\end{tabular}}
\end{table*}

%

\subsection{Feature-Shift Data Heterogeneity}\label{app:results:feature_shift}

\Cref{tab:feature_shift} illustrates the evaluation
under feature shift data heterogeneity among clients.
Unlike convention classification tasks,
this scenarios raise additional
challenges for handling data heterogeneity
from diverse domains.
It can be observed that on all domains
the accuracy improvements over the CLIP baseline
after collaborative training,
are not as signification as federated prompt learning
from a regular dataset without such domain gaps~\Cref{tab:global}.
Note that all methods fail to generalize
on the \textbf{Quickdraw} dataset
because this dataset contains abstract images
that is very challenging even for human recognition.
\begin{table*}[ht]
\centering\caption{%
    \textbf{Comparison of domain-specific accuracy (\%)
    under feature shifts data heterogenity.}
    The \textbf{Avg.}\ column
    denotes a weighted average of the accuracies of all domains
    based on their corresponding image counts.
}\label{tab:feature_shift}
\adjustbox{max width=\linewidth}{%
\begin{tabular}{l|cccccc|c|c}
    \toprule
    Feature Shift
        & \textbf{Clipart} & \textbf{Infograph} & \textbf{Painting}
        & \textbf{Quickdraw} & \textbf{Real} & \textbf{Sketch}
        & \textbf{Avg.} & \textbf{\#} \\
    \midrule
    CLIP      & 54.8  & 40.9  & 48.8   & 6.0  & 77.7  & 49.3    & 44.6    & - \\
    \midrule
    PromptFL
        & \tpm{59.6}{0.2}  & \tpm{45.6}{0.3}  & \tupm{53.7}{0.3}
        & \tpm{8.9}{0.1}   & \tpm{79.8}{0.1}  & \tbpm{54.2}{0.2}
        & \tupm{48.1}{0.1}  & - \\
    FedOTP
        & \tpm{58.4}{0.2}  & \tpm{45.2}{0.1}  & \tpm{53.4}{0.2}
        & \tupm{9.0}{0.1}   & \tpm{79.2}{0.1}  & \tpm{53.2}{0.1}
        & \tpm{47.7}{0.1}  & 1 \\
    \midrule
    \fed{CoCoOp}
        & \tbpm{60.0}{0.1}  & \tbpm{46.1}{0.2}  & \tpm{53.0}{0.3}
        & \tbpm{9.1}{0.2}   & \tpm{79.8}{0.2}  & \tbpm{54.2}{0.2}
        & \tupm{48.1}{0.2}  & \textbf{5} \\
    \fed{PLOT}
        & \tpm{58.5}{0.3}  & \tpm{44.8}{0.2}  & \tpm{53.0}{0.1}
        & \tupm{9.0}{0.4}   & \tpm{79.2}{0.1}  & \tpm{53.3}{0.1}
        & \tpm{47.6}{0.1}  & 1 \\
    \fed{ProDA}
        & \tpm{59.5}{0.2}  & \tpm{45.6}{0.1}  & \tbpm{53.8}{0.2}
        & \tupm{9.0}{0.2}   & \tpm{79.6}{0.1}  & \tpm{54.0}{0.3}
        & \tpm{48.0}{0.1}  & 3 \\
    \fed{ProGrad}
        & \tpm{58.8}{0.2}  & \tpm{44.5}{0.2}  & \tpm{52.5}{0.1}
        & \tpm{7.5}{0.2}   & \tupm{80.0}{0.1}  & \tpm{53.0}{0.0}
        & \tpm{47.3}{0.1}  & 1 \\
    \fed{SRC}
        & \tpm{59.0}{0.1}  & \tpm{44.6}{0.4}  & \tpm{52.6}{0.1}
        & \tpm{7.8}{0.1}   & \tpm{79.7}{0.1}  & \tpm{52.9}{0.1}
        & \tpm{47.3}{0.1}  & 0 \\
    \fed{KgCoOp}
        & \tupm{59.9}{0.1}  & \tupm{45.9}{0.2}  & \tpm{53.6}{0.1}
        & \tpm{8.8}{0.1}   & \tbpm{80.2}{0.1}  & \tupm{54.1}{0.1}
        & \tbpm{48.2}{0.0}  & \underline{4} \\
    \bottomrule
\end{tabular}}
\end{table*}

\subsection{Cross-domain Generalization}\label{app:results:cross}

In~\Cref{tab:xdomain},
we evaluated the cross domain generalization
by training the clients on ImageNet
and evaluating the performance
of the produced global model
on test sets that present domain shift,
including ImageNet-A, ImageNet-R, and ImageNet-S.
On this challenging task,
\fed{CoCoOp} and \fed{KgCoOp},
give best results on a majority of datasets.
\fed{CoCoOp} introduces test-time image
features to generate the prompt.
This reduce the domain gap between
training and testing images.
\fed{KgCoOp} leverages diverse prompt templates
for self-regularization
to maintain the performance
under domain shift.
This could be credited to the rich semantic clues
resilient to diverse domain shifts
in these predefined prompt templates.
\begin{table}[t]
\centering\caption{%
    Comparing the \textbf{cross-domain performance}
    of federated prompt learning methods.
    Here,
    the source domain is ImageNet (IN),
    and the target domains are ImageNet-A(dversarial), -R(endition), and -S(ketch),
    respectively denoted as IN-A, IN-R, IN-S.
}\label{tab:xdomain}
\adjustbox{max width=\linewidth}{%
\begin{tabular}{l|ccc|cc}
    \toprule
    Cross-domain \textaccxd{IN}{\( \cdot \)}
        & \textbf{IN-A} & \textbf{IN-R} & \textbf{IN-S} & Avg.  & \# \\
    \midrule
    CLIP
        & 21.7   & 56.1   & 33.4
        & 37.1   & - \\
    \midrule
    PromptFL
        & \tbpm{24.9}{0.4}   & \tpm{58.2}{0.3}   & \tpm{35.6}{0.6}
        & 39.6   & - \\
    FedOTP
        & \tpm{23.8}{0.3}   & \tpm{58.3}{0.8}   & \tpm{35.2}{0.4}
        & 39.1   & \underline{1} \\
    \midrule
    \fed{CoCoOp}
        & \tpm{24.0}{0.6}   & \tbpm{59.8}{1.1}   & \tbpm{36.0}{1.0}
        & \textbf{39.9}   & \textbf{2}\\
    \fed{PLOT}
        & \tpm{23.8}{0.7}   & \tpm{57.5}{0.3}   & \tpm{34.6}{0.6}
        & 38.6   & 0\\
    \fed{ProDA}
        & \tupm{24.7}{1.7}   & \tpm{58.3}{0.6}   & \tpm{35.6}{0.8}
        & 39.5   & \underline{1}\\
    \fed{ProGrad}
        & \tpm{23.5}{1.1}   & \tpm{58.3}{0.5}   & \tpm{35.5}{1.2}
        & 39.1   & \underline{1}\\
    \fed{SRC}
        & \tpm{24.0}{0.7}   & \tupm{58.7}{0.9}   & \tpm{35.1}{0.5}
        & 39.3   & \underline{1}\\
    \fed{KgCoOp}
        & \tupm{24.7}{1.2}   & \tupm{58.7}{1.4}   & \tupm{35.9}{0.7}
        & \underline{39.8}   & \textbf{2}\\
    \bottomrule
\end{tabular}}
\end{table}

Remarkably,
we reveal that even CLIP models are pretrained
with rich features from large-scale data,
generalization on dataset with large domain shifts
is still challenging for all methods.
For instance,
on the ImageNet-A and ImageNet-S datasets,
all federated prompt learning methods
can only achieve relatively low accuracy (around 20\% to 35\%)
compared with the original ImageNet testing accuracy (73.3\%).
While this may hint at the lack of pretraining of CLIP
on sketch and adversarial images,
it also necessitates future research endeavors
for improving the generalization ability
and robustness
of these federated prompt learning methods.

\subsection{Cost-performance Trade-offs}\label{app:results:cost}

\Cref{tab:cost:prompts,tab:cost:tokens}
present the communication and performance trade-off
by changing the number of prompts or prompt context token length.
\begin{table*}[t]
\centering\caption{%
    \textbf{Trade-offs between accuracy (\%)
    and the number of communicated parameters (in millions)
    under different number of prompts
    on Caltech.}
    Here,
    we sweep the number of prompts
    with \( \braces{1,2,4} \)
    while keeping the number of prompt tokens fixed at 4.
}\label{tab:cost:prompts}
\adjustbox{max width=\linewidth}{%
\begin{tabular}{l|cr|cr|cr|c|c}
    \toprule
    Number
        & \multicolumn{2}{c|}{\textbf{1}}
        & \multicolumn{2}{c|}{\textbf{2}}
        & \multicolumn{2}{c|}{\textbf{4}}
        & \textbf{Avg.} & \textbf{\#} \\
    of Prompts
        & Accuracy & Cost
        & Accuracy & Cost
        & Accuracy & Cost
        & & \\
    \midrule
    PromptFL
        & \tpm{91.5}{0.5} & 2.05
        & \tpm{91.4}{0.4} & 4.10
        & \tpm{91.7}{0.0} & 8.19
        & 91.6  & - \\
    FedOTP
        & \tupm{91.8}{0.1} & 4.10
        & \tupm{91.8}{0.5} & 8.19
        & \tupm{91.9}{0.3} & 16.38
        & \underline{91.8}  & \textbf{3}\\
    \midrule
    \fed{CoCoOp}
        & \tpm{91.7}{0.3} &100.93
        & \tpm{91.3}{0.3} & 102.98
        & \tpm{91.7}{0.0} & 107.07
        & 91.5  & 1 \\
    \fed{PLOT}
        & \tpm{91.6}{0.3} & 2.05
        & \tpm{91.2}{0.3} & 4.10
        & \tpm{91.4}{0.2} & 8.19
        & 91.4  & 1 \\
    \fed{ProDA}
        & \tpm{91.6}{0.3} & 4.10
        & \tpm{91.1}{0.1} & 8.19
        & \tpm{91.7}{0.1} & 16.38
        & 91.5  & 1\\
    \fed{ProGrad}
        & \tpm{90.7}{0.2} & 2.05
        & \tpm{91.1}{0.1} & 4.10
        & \tpm{91.4}{0.1} & 8.19
        & 91.1  & 0\\
    \fed{SRC}
        & \tbpm{92.0}{0.8} & 2.05
        & \tbpm{92.0}{0.3} & 4.10
        & \tbpm{92.1}{0.1} & 8.19
        & \textbf{92.0}  & \textbf{3} \\
    \fed{KgCoOp}
        & \tupm{91.8}{0.2} & 2.05
        & \tpm{91.4}{0.2} & 4.10
        & \tupm{91.9}{0.2} & 8.19
        & 91.7  & \underline{2} \\
    \bottomrule
\end{tabular}}
\end{table*}

\begin{table*}[t]
\centering\caption{%
    \textbf{Trade-offs between accuracy (\%)
    and the number of communicated parameters (in millions)
    under different number of prompt tokens
    on Caltech.}
    Here,
    we sweep the \textbf{number of tokens}
    with \( \braces{4,8,16} \)
    while keeping the number of prompts fixed at 1.
}\label{tab:cost:tokens}
\adjustbox{max width=\linewidth}{%
\begin{tabular}{l|cr|cr|cr|c|c}
    \toprule
    Number
        & \multicolumn{2}{c|}{\textbf{4}}
        & \multicolumn{2}{c|}{\textbf{8}}
        & \multicolumn{2}{c|}{\textbf{16}}
        & \textbf{Avg.} & \textbf{\#} \\
    of Tokens
        & Accuracy & Cost
        & Accuracy & Cost
        & Accuracy & Cost
        & & \\
    \midrule
    PromptFL
        & \tpm{91.5}{0.5} & 2.05
        & \tpm{91.0}{0.6} & 4.10
        & \tpm{91.7}{0.2} & 8.19
        & 91.4 & - \\
    FedOTP
        & \tupm{91.8}{0.1} & 4.10
        & \tupm{91.8}{0.2} & 8.19
        & \tupm{92.0}{0.3} & 16.38
        & \underline{91.8} & \textbf{3}  \\
    \midrule
    \fed{CoCoOp}
        & \tpm{91.7}{0.3} & 100.93
        & \tpm{91.6}{0.9} & 102.98
        & \tpm{91.8}{0.2} & 107.07
        & 91.7 & \textbf{3}  \\
    \fed{PLOT}
        & \tpm{91.6}{0.3} & 2.05
        & \tpm{91.4}{0.3} & 4.10
        & \tpm{91.7}{0.0} & 8.19
        & 91.6 & \underline{2}  \\
    \fed{ProDA}
        & \tpm{91.6}{0.3} & 4.10
        & \tpm{91.5}{0.6} & 8.19
        & \tpm{91.8}{0.2} & 16.38
        & 91.6 & \textbf{3}  \\
    \fed{ProGrad}
        & \tpm{90.7}{0.2} & 2.05
        & \tpm{91.0}{0.3} & 4.10
        & \tpm{91.6}{0.1} & 8.19
        & 91.1 & 1  \\
    \fed{SRC}
        & \tbpm{92.0}{0.8} & 2.05
        & \tbpm{92.1}{0.3} & 4.10
        & \tbpm{92.2}{0.1} & 8.19
        & \textbf{92.1} & \textbf{3}  \\
    \fed{KgCoOp}
        & \tupm{91.8}{0.2} & 2.05
        & \tpm{91.2}{0.4} & 4.10
        & \tpm{91.4}{0.0} & 8.19
        & 91.5 & \underline{2}  \\
    \bottomrule
\end{tabular}}
\end{table*}

%

\subsection{Evaluation on Transformer Image Encoder}

In~\Cref{%
    tab:global_vit,%
    tab:personal_vit,%
    tab:base2novel_random_vit},
we respectively report the global, personal
and base-to-novel accuracy metrics
for the ViT-B/16 image encoder,
following the evaluation protocols
used in \Cref{tab:global,tab:personal,tab:base2novel_random}.
To sum up,
these results show a similar trend
on those of ResNet-50,
and the ViT-B/16 in most experiments delivers
better results than the ResNet-50 image encoder.
\begin{table*}[ht]
    \centering
    \caption{%
        \textbf{%
        Comparison of shared global model accuracy
        \textaccglobal{} (\%)
        of federated prompt learning methods
        with a ViT-B/16 image encoder.}
        Results are reported
        in a similar style as \Cref{tab:global}.
    }\label{tab:global_vit}
    \adjustbox{max width=\linewidth}{%
    \begin{tabular}{l|rrrrrrrr|r|c}
        \toprule
        \multicolumn{1}{c|}{Global \textaccglobal{}}
            & \theader{Caltech} & \theader{DTD}
            & \theader{Aircraft} & \theader{Food}
            & \theader{Cars} & \theader{Flowers}
            & \theader{Pets} & \theader{UCF}
            & \multicolumn{1}{|c|}{\textbf{Avg.}} & \theader{\#} \\
        \midrule
        \textbf{ZS-CLIP}
            & \tpm{93.5}{0.0}  & \tpm{45.0}{0.0}
            & \tpm{24.3}{0.0}  & \tpm{85.5}{0.0}
            & \tpm{65.6}{0.0}  & \tpm{68.0}{0.0}
            & \tpm{89.2}{0.0}  & \tpm{67.5}{0.0}
            & 67.3   & - \\
        \midrule
        \textbf{PromptFL}
            & \tupm{95.5}{0.1}  & \tupm{59.6}{0.7}
            & \tpm{31.2}{0.3}  & \tpm{86.8}{0.1}
            & \tpm{70.2}{1.3}  & \tbpm{86.2}{1.7}
            & \tpm{92.4}{0.2}  & \tbpm{77.5}{0.6}
            & \underline{74.9}   & - \\
        \textbf{FedOTP}
            & \tbpm{95.6}{0.1}  & \tbpm{60.6}{0.7}
            & \tbpm{32.4}{0.6}  & \tupm{86.9}{0.2}
            & \tupm{70.6}{0.5}  & \tpm{85.6}{0.3}
            & \tpm{92.2}{0.5}  & \tpm{76.6}{0.6}
            & \textbf{75.1}   & 5 \\
        \midrule
        \textbf{\fed{CoCoOp}}
            & \tbpm{95.6}{0.2}  & \tpm{57.7}{1.0}
            & \tpm{31.0}{0.6}  & \tpm{86.6}{0.2}
            & \tpm{68.5}{0.3}  & \tpm{81.6}{0.8}
            & \tpm{92.4}{0.7}  & \tpm{74.5}{0.5}
            & 73.5   & 1 \\
        \textbf{\fed{PLOT}}
            & \tpm{95.4}{0.2}  & \tupm{59.6}{0.9}
            & \tpm{31.3}{0.4}  & \tpm{86.5}{0.1}
            & \tpm{70.1}{1.3}  & \tupm{85.8}{2.2}
            & \tpm{92.4}{0.2}  & \tbpm{77.5}{0.3}
            & 74.8   & 3 \\
        \textbf{\fed{ProDA}}
            & \tpm{95.4}{0.2}  & \tpm{58.6}{1.0}
            & \tpm{31.3}{0.8}  & \tpm{86.6}{0.1}
            & \tbpm{70.8}{1.2}  & \tpm{84.7}{0.4}
            & \tupm{92.5}{0.2}  & \tupm{77.0}{0.5}
            & 74.6   & 3 \\
        \textbf{\fed{ProGrad}}
            & \tpm{95.2}{0.1}  & \tpm{56.5}{0.4}
            & \tpm{30.0}{0.4}  & \tbpm{87.1}{0.1}
            & \tpm{69.3}{0.1}  & \tpm{81.3}{1.4}
            & \tbpm{92.6}{0.3}  & \tpm{75.4}{0.5}
            & 73.4   & 2 \\
        \textbf{\fed{PromptSRC}}
            & \tpm{94.0}{0.5}  & \tpm{58.1}{0.3}
            & \tpm{31.4}{0.2}  & \tpm{86.8}{0.1}
            & \tpm{70.3}{0.2}  & \tpm{85.3}{0.6}
            & \tupm{92.5}{0.2}  & \tpm{75.1}{0.1}
            & 74.2   & 3 \\
        \textbf{\fed{KgCoOp}}
            & \tpm{95.4}{0.1}  & \tpm{59.4}{0.6}
            & \tupm{31.6}{0.5}  & \tupm{86.9}{0.0}
            & \tpm{70.2}{1.0}  & \tpm{83.7}{1.5}
            & \tupm{92.5}{0.4}  & \tpm{76.7}{0.2}
            & 74.6   & 4 \\
        \bottomrule
    \end{tabular}}
\end{table*}

\begin{table*}[ht]
    \centering
    \caption{%
        \textbf{%
        Comparison of personal model accuracy
        \textaccpersonal{} (\%)
        of federated prompt learning methods
        on various datasets
        with a ViT-B/16 image encoder.}
    }\label{tab:personal_vit}
    \adjustbox{max width=\linewidth}{%
    \begin{tabular}{l|rrrrrrrr|r|c}
        \toprule
        \multicolumn{1}{c|}{Personal \textaccpersonal{}}
            & \theader{Caltech} & \theader{DTD}
            & \theader{Aircraft} & \theader{Food}
            & \theader{Cars} & \theader{Flowers}
            & \theader{Pets} & \theader{UCF}
            & \multicolumn{1}{|c|}{\textbf{Avg.}} & \theader{\#} \\
        \midrule
        \textbf{ZS-CLIP}
            & \tpm{93.5}{0.0}  & \tpm{45.0}{0.0}
            & \tpm{24.3}{0.0}  & \tpm{85.5}{0.0}
            & \tpm{65.6}{0.0}  & \tpm{68.0}{0.0}
            & \tpm{89.2}{0.0}  & \tpm{67.5}{0.0}
            & 67.3  & - \\
        \midrule
        \textbf{PromptFL}
            & \tpm{95.7}{0.4}  & \tupm{73.3}{0.6}
            & \tpm{43.6}{0.5}  & \tupm{89.1}{0.6}
            & \tpm{76.7}{1.3}  & \tpm{88.5}{1.0}
            & \tpm{92.8}{1.1}  & \tpm{82.5}{0.6}
            & 80.3  & - \\
        \textbf{FedOTP}
            & \tbpm{96.2}{0.5}  & \tbpm{75.2}{2.2}
            & \tbpm{46.8}{1.1}  & \tbpm{90.1}{0.9}
            & \tbpm{77.7}{1.8}  & \tbpm{91.6}{0.2}
            & \tupm{93.1}{0.7}  & \tbpm{84.2}{1.5}
            & \textbf{81.9}  & \textbf{8} \\
        \midrule
        \textbf{\fed{CoCoOp}}
            & \tupm{95.8}{0.5}  & \tpm{72.2}{1.7}
            & \tupm{45.9}{0.1}  & \tbpm{90.1}{0.8}
            & \tpm{75.8}{0.6}  & \tpm{88.1}{0.6}
            & \tbpm{93.9}{0.7}  & \tupm{82.7}{1.1}
            & \underline{80.6}  & \underline{5} \\
        \textbf{\fed{PLOT}}
            & \tupm{95.8}{0.3}  & \tpm{71.5}{0.3}
            & \tpm{44.3}{1.0}  & \tpm{88.7}{0.4}
            & \tpm{76.8}{1.3}  & \tpm{88.8}{1.8}
            & \tpm{92.9}{1.2}  & \tpm{82.0}{1.2}
            & 80.1  & \underline{5} \\
        \textbf{\fed{ProDA}}
            & \tupm{95.8}{0.4}  & \tpm{70.3}{1.1}
            & \tpm{43.3}{1.4}  & \tpm{88.7}{0.4}
            & \tupm{76.9}{1.7}  & \tupm{89.1}{2.0}
            & \tpm{92.9}{0.5}  & \tpm{82.0}{0.6}
            & 79.9  & 4 \\
        \textbf{\fed{ProGrad}}
            & \tpm{95.1}{0.3}  & \tpm{70.2}{0.7}
            & \tpm{44.4}{1.2}  & \tpm{88.9}{0.6}
            & \tpm{75.9}{1.0}  & \tpm{87.3}{1.5}
            & \tpm{92.3}{1.0}  & \tpm{81.5}{0.8}
            & 79.5  & 1 \\
        \textbf{\fed{PromptSRC}}
            & \tpm{94.8}{0.5}  & \tpm{71.2}{1.4}
            & \tpm{44.2}{0.8}  & \tpm{88.4}{0.8}
            & \tpm{76.8}{0.7}  & \tpm{88.0}{0.9}
            & \tpm{92.7}{1.4}  & \tpm{82.4}{0.8}
            & 79.8  & 2 \\
        \textbf{\fed{KgCoOp}}
            & \tpm{95.5}{0.2}  & \tpm{71.4}{1.4}
            & \tpm{40.4}{0.7}  & \tpm{88.7}{1.0}
            & \tpm{75.3}{1.3}  & \tpm{88.0}{1.1}
            & \tpm{92.2}{0.7}  & \tpm{81.2}{1.8}
            & 79.1  & 0 \\
        \bottomrule
    \end{tabular}}
    \cvspace{-10pt}
\end{table*}

\begin{table*}[!ht]
    \centering\caption{%
        \textbf{Comparison of base and novel class accuracy \((\%)\)
        of federated prompt learning methods
        with a ViT-B/16 image encoder.}
        Evaluation follows \Cref{tab:base2novel_random}
        except the use of a ViT-B/16 image encoder.
    }\label{tab:base2novel_random_vit}
    \cvspace{-0.5em}
    \setlength{\tabcolsep}{2.5pt}
    \adjustbox{max width=\linewidth}{%
    \begin{tabular}{l|ccc|ccc|ccc|ccc|ccc|c}
        \multicolumn{1}{c}{}
            & \multicolumn{3}{c}{\textbf{Caltech}}
            & \multicolumn{3}{c}{\textbf{Aircraft}}
            & \multicolumn{3}{c}{\textbf{Cars}}
            & \multicolumn{3}{c}{\textbf{Flowers}}
            & \multicolumn{3}{c}{\textbf{Avg.}}
            & \\
        \toprule
        Metric
            & \textaccbase & \textaccnovel & \textaccharmonic
            & \textaccbase & \textaccnovel & \textaccharmonic
            & \textaccbase & \textaccnovel & \textaccharmonic
            & \textaccbase & \textaccnovel & \textaccharmonic
            & \textaccbase & \textaccnovel & \textaccharmonic
            & \theader{\#} \\
        \midrule
        \textbf{ZS-CLIP}
            & 95.6  & 95.5  & 95.5
            & 29.5  & 34.1  & 31.6
            & 67.1  & 76.5  & 71.5
            & 81.6  & 68.0  & 74.2
            & \txtg{68.4} & \txtg{68.5}  & 68.2 & -\\
        \midrule
        \textbf{PromptFL}
            & \tgpm{96.6}{0.3} & \tgpm{95.6}{0.4}  & \tpm{96.1}{0.3}
            & \tgpm{32.2}{1.2} & \tgpm{34.6}{1.0}  & \tpm{33.4}{1.1}
            & \tgpm{73.2}{0.8} & \tgpm{74.9}{0.8}  & \tpm{74.0}{0.3}
            & \tgpm{87.4}{0.1} & \tgpm{69.1}{1.1}  & \tpm{77.2}{0.7}
            & \txtg{72.4} & \txtg{68.5}  & 70.2   & -\\
        \textbf{FedOTP}
            & \tgpm{97.3}{0.2} & \tgpm{95.4}{0.8}  & \tupm{96.3}{0.4}
            & \tgpm{32.5}{1.8} & \tgpm{34.7}{2.2}  & \tpm{33.5}{0.5}
            & \tgpm{72.1}{0.3} & \tgpm{74.5}{0.1}  & \tpm{73.3}{0.1}
            & \tgpm{86.5}{2.5} & \tgpm{70.5}{0.8}  & \tpm{77.6}{0.6}
            & \txtg{72.1} & \txtg{68.8}  & 70.2   & \underline{3}  \\
        \midrule
        \textbf{\fed{CoCoOp}}
            & \tgpm{96.5}{0.2} & \tgpm{96.0}{0.6}  & \tpm{96.2}{0.3}
            & \tgpm{30.4}{2.6} & \tgpm{35.8}{1.0}  & \tpm{32.9}{2.6}
            & \tgpm{71.7}{0.4} & \tgpm{75.3}{0.3}  & \tpm{73.5}{0.4}
            & \tgpm{84.0}{1.5} & \tgpm{72.1}{1.0}  & \tpm{77.6}{0.2}
            & \txtg{69.7} & \txtg{67.3}  & 68.1   & 2  \\
        \textbf{\fed{PLOT}}
            & \tgpm{96.5}{0.4} & \tgpm{95.7}{0.3}  & \tpm{96.1}{0.3}
            & \tgpm{32.7}{0.8} & \tgpm{34.7}{0.7}  & \tupm{33.7}{0.8}
            & \tgpm{71.7}{0.0} & \tgpm{76.2}{0.0}  & \tpm{73.9}{0.0}
            & \tgpm{88.4}{4.3} & \tgpm{68.6}{2.5}  & \tpm{77.2}{1.2}
            & \txtg{54.4} & \txtg{49.8}  & 51.7   & 2  \\
        \textbf{\fed{ProDA}}
            & \tgpm{96.7}{0.4} & \tgpm{95.1}{1.1}  & \tpm{95.9}{0.6}
            & \tgpm{31.7}{1.1} & \tgpm{35.7}{1.1}  & \tpm{33.5}{0.8}
            & \tgpm{72.6}{0.6} & \tgpm{74.7}{0.5}  & \tpm{73.6}{0.1}
            & \tgpm{86.0}{2.5} & \tgpm{70.3}{1.5}  & \tpm{77.3}{0.7}
            & \txtg{71.7} & \txtg{69.0}  & 70.1   & 2  \\
        \textbf{\fed{ProGrad}}
            & \tgpm{96.8}{0.3} & \tgpm{96.2}{0.4}  & \tbpm{96.5}{0.4}
            & \tgpm{32.4}{0.5} & \tgpm{34.4}{1.2}  & \tpm{33.4}{0.8}
            & \tgpm{72.0}{0.5} & \tgpm{76.5}{0.6}  & \tpm{74.2}{0.1}
            & \tgpm{86.6}{2.0} & \tgpm{70.3}{0.8}  & \tpm{77.6}{0.7}
            & \txtg{72.0} & \txtg{69.4}  & 70.4   & \textbf{4}  \\
        \textbf{\fed{SRC}}
            & \tgpm{96.7}{0.1} & \tgpm{95.8}{0.2}  & \tpm{96.2}{0.0}
            & \tgpm{32.2}{1.0} & \tgpm{35.5}{0.8}  & \tbpm{33.8}{0.7}
            & \tgpm{72.4}{0.3} & \tgpm{77.0}{0.2}  & \tbpm{74.6}{0.1}
            & \tgpm{86.4}{0.5} & \tgpm{73.4}{0.5}  & \tbpm{79.4}{0.5}
            & \txtg{71.9} & \txtg{70.4}  & \textbf{71.0}   & \textbf{4}  \\
        \textbf{\fed{KgCoOp}}
            & \tgpm{96.7}{0.6} & \tgpm{96.0}{0.2}  & \tupm{96.3}{0.3}
            & \tgpm{33.4}{0.5} & \tgpm{34.3}{1.0}  & \tbpm{33.8}{0.6}
            & \tgpm{72.9}{1.0} & \tgpm{75.9}{0.2}  & \tupm{74.3}{0.5}
            & \tgpm{88.0}{2.1} & \tgpm{70.6}{0.4}  & \tupm{78.3}{0.7}
            & \txtg{72.8} & \txtg{69.2}  & \underline{70.7}   & \textbf{4}  \\
        \bottomrule
    \end{tabular}}
        \vspace{-10pt}
    \end{table*}

\section{Discussion}\label{app:discussion}

\subsection{Implementation Details}\label{app:discussion:implementation}
\textbf{Environments}
We implement all evaluated methods
with PyTorch~\cite{paszke2017automatic}
of version 2.1.0.
We try to minimize the number of packages
used in our code framework,
and setting up the environment only requires minutes.
To alleviate computational burden,
we applied the automatic mixed precision (AMP) training%
\footnote{\url{https://pytorch.org/docs/stable/amp.html}.},
which leverages the 16-bit floating point format
to reduce GPU memory consumption and computation cost.

\textbf{Code Framework}
To date,
there still lacks a comprehensive and reliable evaluation
of federated prompt learning algorithms for vision tasks.
Zhou~\etal{}~\cite{zhou2022cocoop}
established the first seminal library
for \emph{centralized} prompt learning,
which is later reused by a line of subsequent works.
However,
this library is not tailored for \emph{federated learning}.
Namely,
it poses additional challenges
to incorporate various federated algorithms
with existing prompt learning techniques
in a flexible and scalable way.

To close this gap,
we release the first framework
with large-scale evaluations
to push the frontier of federated prompt learning.
To harvest the rapid progress
from federated learning and prompt learning literature,
we decouple the design
of the federated learning and prompt learning modules,
making it easier to integrate the progress
from both research fields
in a scalable and efficient way.
We simplify and unify the interface of data-loading
to achieve better adaptation of new datasets
and also make it readily available for users
to adapt to their customized datasets for new tasks
with minimal modification.
We plan to actively support more applications
beyond the evaluated image classification tasks.

\subsection{Computational Resources}\label{app:discussion:resources}

The experiments are conducted
on a cluster consisting of multiple servers
equipped with NVIDIA A100 graphic cards.
We run experiments on servers
equipped with the SLURM%
\footnote{\url{https://slurm.schedmd.com/documentation.html}.}
job scheduler.

\end{document}